\documentclass[]{imag-ms-template}

\usepackage{graphicx}
\usepackage{subcaption}
\usepackage{float}
\usepackage{multirow}
\usepackage{bm}
\usepackage{amsmath}
\usepackage{lineno}
\usepackage{xurl}

\newcommand{\suppletter}{A}

\newlength{\subfigheight}

\DeclareSourcemap{
  \maps[datatype=bibtex, overwrite]{
    \map{\step[fieldset=abstract, null]}
    \map{\step[fieldset=file, null]}
    \map{\step[fieldset=keywords, null]}
    \map{\step[fieldset=annote, null]}
  }
}

\title{Brain2Vec: A Self-Supervised 3D Foundation Model for Structural Brain MRI}

\author{Spiros Maggioros,$^{1\ast}$ Guray Erus,$^{1}$ Gareth Harman,$^{1}$\\
George Aidinis,$^{1}$ Alexander Getka,$^{1}$ Aristeidis Sotiras,$^{1}$ Christos Davatzikos$^{1}$\\
{\small $^{1}$Center for Biomedical Image Computing and Analytics (CBICA) }\\
{\small Perelman School of Medicine, University of Pennsylvania, Philadelphia, PA, USA}\\
{\small $^\ast$Correspondence: Spiros.Maggioros@pennmedicine.upenn.edu}
}

\begin{document}

\maketitle


\begin{abstract}
Deep learning models have become essential tools in medical imaging, yet training them end-to-end for diverse tasks remains challenging due to limited labeled data and computational constraints. To address this, we introduce Brain2Vec, a foundation model pre-trained on 74,425 T1-weighted Magnetic Resonance Imaging (MRI) scans pooled from multiple studies. Brain2Vec can serve both as a weight initialization backbone for fine-tuning downstream tasks and as a standalone feature extractor. We evaluated Brain2Vec in three tasks: downstream predictive performance, convergence speed, and embedding quality. Using Brain2Vec's pretrained weights as initialization and fine-tuning on labeled data for a brain age prediction task, results demonstrate improved final accuracy and faster convergence compared to random initialization. Additionally, its frozen embeddings outperform anatomically defined volumetric features on four of five downstream classification tasks and perform competitively with a state-of-the-art foundation model pre-trained on substantially more data. On the independent BrainFMBench benchmark, Brain2Vec ranked first among six pretrained brain MRI foundation models on 9 of 15 sex, age, and BMI prediction tasks across five cohorts. Lastly, we show that Brain2Vec encodes information primarily from a spatially localized and neuroanatomically meaningful set of brain regions, and that manipulating a subset of latent features produces monotonic, region-specific effects on the reconstructed anatomy. Brain2Vec code and trained models are available as an open-source package on \url{https://github.com/CBICA/Brain2Vec}. Users can apply Brain2Vec locally by installing the package, or directly through the NiChart cloud platform \url{https://cloud.neuroimagingchart.com}, where Brain2Vec features can be derived without requiring local installation or specialized infrastructure.
\end{abstract}

\section{Introduction}

Traditional machine learning approaches for deriving imaging biomarkers from structural Magnetic Resonance Imaging (MRI) typically rely on a predefined feature extraction step, such as segmentation or parcellation of regional anatomy, followed by supervised modeling on the resulting measurements \citep{doshi_muse_2016, bashyam_dlmuse_2025}. These pipelines have supported a wide range of neuroimaging applications, including brain age estimation and disease prediction \citep{Rathore2017}, but they depend on predefined anatomical segmentations and task-specific model development.

Deep learning (DL) has made it possible to learn predictive representations directly from images, potentially bypassing hand-engineered feature pipelines. In practice, however, training end-to-end models for specific neuroimaging tasks poses significant challenges. MRI volumes are substantially more computationally demanding than 2D natural images, with 3D volume-based training incurring considerable memory and computational costs \citep{chen_masked_2023, 10230477}. At the same time, neuroimaging datasets are typically small, labeled datasets are limited, and class distributions are often imbalanced, contrasting sharply with the large-scale labeled datasets available in computer vision, such as ImageNet \citep{5206848} with over 14 million images for 2D classification. These constraints hinder the development of robust task-specific deep models and motivate the need for transferable representations learned from large, heterogeneous MRI collections.

Self-supervised pre-training offers a practical way to mitigate these limitations by learning transferable image representations from large collections of unlabeled scans \citep{huang_self-supervised_2023, he_masked_2022}. Rather than optimizing directly for a single downstream target, self-supervised models are trained using a generic pretext objective and can then be adapted to classification or regression tasks using substantially smaller labeled datasets. This paradigm is especially appealing in neuroimaging, where large-scale imaging repositories are increasingly available, but task-specific annotations remain expensive, heterogeneous, and often scarce.

In this work, we present Brain2Vec, a self-supervised foundation model pre-trained on 74,425 T1-weighted MRI scans pooled from multiple studies within the iSTAGING harmonized dataset \citep{habes_brain_2021}. The training data span a broad range of scanner hardware, acquisition protocols, and clinical populations, allowing the model to learn a high-dimensional representation of neuroanatomy from diverse real-world data. Brain2Vec produces a 768-dimensional embedding for each scan, which can be used in two complementary ways: as initialization for downstream fine-tuning, or as a frozen feature representation for supervised prediction.

Validation experiments demonstrate that Brain2Vec improves both predictive performance and convergence speed when used for encoder fine-tuning. When evaluated as a feature extractor, Brain2Vec outperformed SPARE-BA, a model trained on volumetric features derived using DLMUSE anatomical segmentation ~\citep{bashyam_dlmuse_2025}, across all tasks except a single AD prediction task, as well as NeuroJEPA \citep{huang2026learningsparselatentpredictive}, a foundation model pre-trained on $1.55$M scans ($683,860$ T1w), on multiple classification and brain age prediction tasks, a particularly notable result given that Brain2Vec was pre-trained on $\approx 9.2\times$ less data. This suggests that learning a strong representation of brain structure may not require web-scale data. Unlike natural-image datasets, where visual content varies enormously across scenes, objects, and viewpoints, preprocessed structural brain MRI is a comparatively low-variance domain: scans are skull-stripped, atlas-registered, and anatomically aligned, so the variability a model must capture is largely confined to fine-grained morphology and acquisition effects. To complement these experiments with an external validation, we also submitted Brain2Vec to BrainFMBench \citep{Encin2026.05.15.725427}, an independently maintained benchmark that compares frozen embeddings from brain MRI foundation models on common tasks and cohorts. There, Brain2Vec was among the strongest models for predicting age, sex, and BMI, despite having been pre-trained on adult cohorts while the benchmark cohorts are largely pediatric and young-adult. Together, these results establish Brain2Vec as an effective self-supervised backbone for structural neuroimaging analysis.

\paragraph{Contributions.}
This work makes the following contributions: 
\begin{enumerate}
    \item First, we introduce Brain2Vec, a self-supervised 3D masked autoencoder trained on 74,425 T1-weighted MRI scans drawn from multiple studies, with the goal of learning transferable representations of brain structure. 
    \item Second, we demonstrate that pretrained Brain2Vec weights serve as a stronger initialization for downstream task-specific applications than random initialization. We illustrate this here through brain age prediction, where we observe improvements in both convergence speed and final accuracy. 
    \item Third, we show that frozen Brain2Vec embeddings support competitive lightweight downstream modeling. They outperform segmentation-based volumetric features on four of five disease and risk-factor classification tasks and perform competitively with NeuroJEPA. In the independent BrainFMBench benchmark, they rank first among pretrained foundation models on most age, sex, and BMI prediction tasks.
    \item Last, we perform an extensive analysis of Brain2Vec latent embeddings, showing that the model encodes information primarily from a spatially localized and neuroanatomically meaningful set of brain regions. We prove that manipulating a subset of latent features produces monotonic, region-specific effects on the reconstructed anatomy, which is important for the interpretability of disease or aging related effects.
\end{enumerate}

\section{Related Work}

Structural neuroimaging has a long tradition of characterizing brain morphology through segmentation-based pipelines, in which anatomically defined regions of interest (ROIs) are delineated and summarized as volumetric, morphometric, or thickness-based features, encompassing measures such as regional volumes, cortical thickness, surface area, and lesion or abnormality loads \citep{fischl_thickness_2000, fischl2012freesurfer, doshi_muse_2016, griffanti_bianca_2016, billot_synthseg_2023}. Imaging features are used in downstream pipelines as inputs to conventional machine learning models for tasks such as disease classification, brain age prediction, and biomarker discovery \citep{habes_advanced_2016, cole_predicting_2017, Rathore2017, hwang_disentangling_2022}. This paradigm has driven substantial progress in neuroimaging research, enabling the discovery of imaging biomarkers and the development of predictive models across a wide range of conditions, including Alzheimer's disease, brain aging, and psychiatric disorders. However, its reliance on a-priori anatomical definitions and task-specific model development inherently constrains its generalizability across diverse clinical applications.

Self-supervised learning has proven particularly well-suited to settings where labeled data are limited but unlabeled scans are abundant. In computer vision, Vision Transformers (ViTs) showed that images can be modeled effectively as sequences of patches, while masked autoencoders (MAEs) demonstrated that reconstructing heavily masked inputs can yield scalable and transferable visual representations \citep{dosovitskiy_image_2020, he_masked_2022}. 

In medical imaging, recent work has highlighted the promise of self-supervised pre-training for improving downstream predictive performance and sample efficiency across classification and segmentation tasks \citep{huang_self-supervised_2023}. Several methods originally developed for computer vision have since been adapted to medical and neuroimaging datasets. Among these, masked autoencoders have demonstrated improvements in medical image classification and segmentation, including in 3D settings where labeled data are particularly costly to obtain \citep{10230477, chen_masked_2023}. Specifically, in brain MRI analysis, domain-tailored self-supervised pre-training has been shown to enhance downstream diagnostic models, and recent work has begun exploring large-scale brain-specific foundation models for neuroimaging tasks \citep{huang_tailored_2025, cox_brainsegfounder_2024, tak_generalizable_2026, huang2026learningsparselatentpredictive}.

\section{Materials and Methods}

\subsection{Datasets}
\label{subsec:datasets}
\label{sec:datasets}

The pre-training dataset comprised 74,425 T1-weighted MRI scans pooled from multiple studies within the iSTAGING consortium \citep{habes_brain_2021}. While the majority of scans were from cognitively unimpaired subjects, the dataset also included individuals with MCI and dementia, enabling the model to capture disease-related neuroanatomical changes. We expect that including diseased groups in pre-training plays a beneficial role in downstream tasks presented in Section~\ref{sec:brain2vec_results} for disease classification. A detailed breakdown of pre-training datasets and sample characteristics is provided in Table~\ref{tab:brain2vec_data_overview}. The dataset was stratified by diagnosis and partitioned into 80\% training, 10\% validation and 10\% test set at the participant level, so that all longitudinal scans from a given participant were assigned to the same set.

\begin{table}[htbp]
\centering
\small
\caption{Overview of datasets used for Brain2Vec pre-training, split by training and validation sets.}
\label{tab:brain2vec_data_overview}
\resizebox{\textwidth}{!}{%
\begin{tabular}{l r c r @{\hspace{1.2cm}} l r c r}
\toprule
\multicolumn{4}{c}{\textit{Training Set}} &
\multicolumn{4}{c}{\textit{Validation Set}} \\
\cmidrule(lr){1-4}\cmidrule(lr){5-8}
\textbf{Study} & \textbf{n} & \textbf{Mean Age $\pm$ Std} & \textbf{(\%) F} &
\textbf{Study} & \textbf{n} & \textbf{Mean Age $\pm$ Std} & \textbf{(\%) F} \\
\midrule
UKBIOBANK  & 32,719 & $64.15 \pm 7.53$  & 52.87 & UKBIOBANK  & 4,104  & $64.26 \pm 7.53$  & 51.32 \\
PENN       & 694    & $73.10 \pm 9.45$  & 57.49 & PENN       & 95     & $73.34 \pm 9.73$  & 55.79 \\
OASIS      & 1,925  & $72.40 \pm 10.57$ & 56.68 & OASIS      & 250    & $72.77 \pm 10.65$ & 57.20 \\
ADNI       & 8,672  & $75.14 \pm 7.38$  & 45.18 & ADNI       & 1,074  & $75.14 \pm 7.41$  & 45.72 \\
AIBL       & 1,499  & $74.21 \pm 6.83$  & 54.30 & AIBL       & 210    & $74.57 \pm 7.00$  & 53.33 \\
CARDIA     & 1,111  & $52.67 \pm 4.30$  & 53.65 & CARDIA     & 142    & $52.45 \pm 4.23$  & 49.30 \\
PreventAD  & 991    & $65.20 \pm 5.56$  & 71.64 & PreventAD  & 114    & $65.13 \pm 5.32$  & 64.04 \\
BLSA       & 3,268  & $73.11 \pm 12.30$ & 52.05 & BLSA       & 377    & $73.55 \pm 12.23$ & 49.34 \\
MESA       & 850    & $74.46 \pm 8.17$  & 52.59 & MESA       & 94     & $73.66 \pm 8.33$  & 58.51 \\
HABS       & 410    & $74.91 \pm 6.37$  & 59.51 & HABS       & 51     & $75.78 \pm 6.00$  & 52.94 \\
HCP        & 1,462  & $41.42 \pm 18.87$ & 55.61 & HCP        & 179    & $41.38 \pm 17.26$ & 51.40 \\
SHIP       & 2,660  & $52.81 \pm 13.67$ & 51.05 & SHIP       & 330    & $53.42 \pm 13.90$ & 51.21 \\
BIOCARD    & 867    & $64.55 \pm 11.08$ & 59.98 & BIOCARD    & 125    & $64.96 \pm 10.18$ & 56.80 \\
GSP        & 1,257  & $21.54 \pm 2.92$  & 58.55 & GSP        & 147    & $21.56 \pm 2.72$  & 53.06 \\
WRAP       & 412    & $65.74 \pm 6.35$  & 66.50 & WRAP       & 52     & $64.83 \pm 6.74$  & 76.92 \\
HANDLS     & 181    & $53.03 \pm 8.96$  & 54.14 & HANDLS     & 24     & $50.02 \pm 10.29$ & 58.33 \\
WHIMS      & 523    & $73.17 \pm 3.93$  & 100.00 & WHIMS      & 67     & $72.38 \pm 3.95$  & 100.00 \\
SPRINT     & 45     & $74.63 \pm 7.24$  & 35.56 & SPRINT     & 6      & $75.95 \pm 3.78$  & 33.33 \\
\midrule
\textbf{Total} & \textbf{59,546 (80\%)} & $\bm{65.00 \pm 12.84}$ & \textbf{52.99} &
\textbf{Total} & \textbf{7,441 (10\%)} & $\bm{65.16 \pm 12.67}$ & \textbf{51.73} \\
\bottomrule
\end{tabular}%
}
\end{table}

For brain age prediction, we used $19,671$ cognitively unimpaired baseline scans from the iSTAGING consortium, spanning 13 studies (Supplementary Table~\ref{tab:ba_data_overview}). To ensure a balanced age distribution across splits, scans were stratified into 10-year age bins and partitioned into 60\% for training, 30\% for validation, and 10\% for testing. For out-of-distribution validation, we used the IXI \citep{ixi_dataset} dataset which consists of $581$ cognitively unimpaired subjects. We removed subjects aged $\leq 40$, resulting in a total of $n = 363$, reflecting the clinical scope of the model where neurodegeneration-related structural changes are most relevant.

For disease classification, we used 21,176 scans from the iSTAGING consortium, spanning 15 studies (Supplementary Tables~\ref{tab:dx_cn_ad}, \ref{tab:dx_cn_diabetes}, \ref{tab:dx_cn_bmi}, \ref{tab:dx_cn_smoking}, and~\ref{tab:dx_cn_hypertension}), covering six diagnostic and risk-factor categories: cognitively unimpaired (CU), Alzheimer's disease (AD), diabetes, elevated BMI, smoking status, and hypertension. To mitigate inter-site variability, data from multiple studies were balanced across disease groups where possible. All classification experiments were evaluated using 5-fold stratified cross-validation. For out-of-distribution validation, we used the MIRIAD \citep{malone2013miriad} dataset, comprising baseline T1 MRI scans from 46 mild-to-moderate Alzheimer's subjects and 23 cognitively normal controls.

All scans underwent the same preprocessing pipeline as with Brain2Vec pre-training introduced in Sec.~\ref{subsec:brain2vec_training}, including skull-stripping, atlas registration, resampling to $128 \times 128 \times 128$ voxels and Z-score normalization.

Our model was also submitted and validated on BrainFMBench \citep{Encin2026.05.15.725427}, where it was compared with five other pretrained foundation models, namely 3D-Neuro-SimCLR \citep{kaczmarek2025buildinggeneralsimclrselfsupervised}, AnatCL \citep{BARBANO2026178} (Global and Local variants), BrainIAC \citep{tak_generalizable_2026}, Dinov3D-neuro, and SwinBrain \citep{suo2025automatic}, as well as with two FreeSurfer-derived feature sets (Schaefer and aparc) and the benchmark maintainers' un-CNN baseline \citep{Encin2026.06.07.730652}, on five total datasets including the Nathan Kline Institute Rockland Sample (958 subjects), Healthy Brain Network (997 subjects), Philadelphia Neurodevelopmental Cohort (1439 subjects), Brazilian High Risk Cohort (542 subjects) and the Chinese Color Nest Project (178 subjects) pooled from the Reproducible Brain Charts (RBC) \citep{shafiei2025reproducible} studies.

\subsection{Architecture}

\begin{figure}[htpb]
    \centering
    \includegraphics[width=\textwidth]{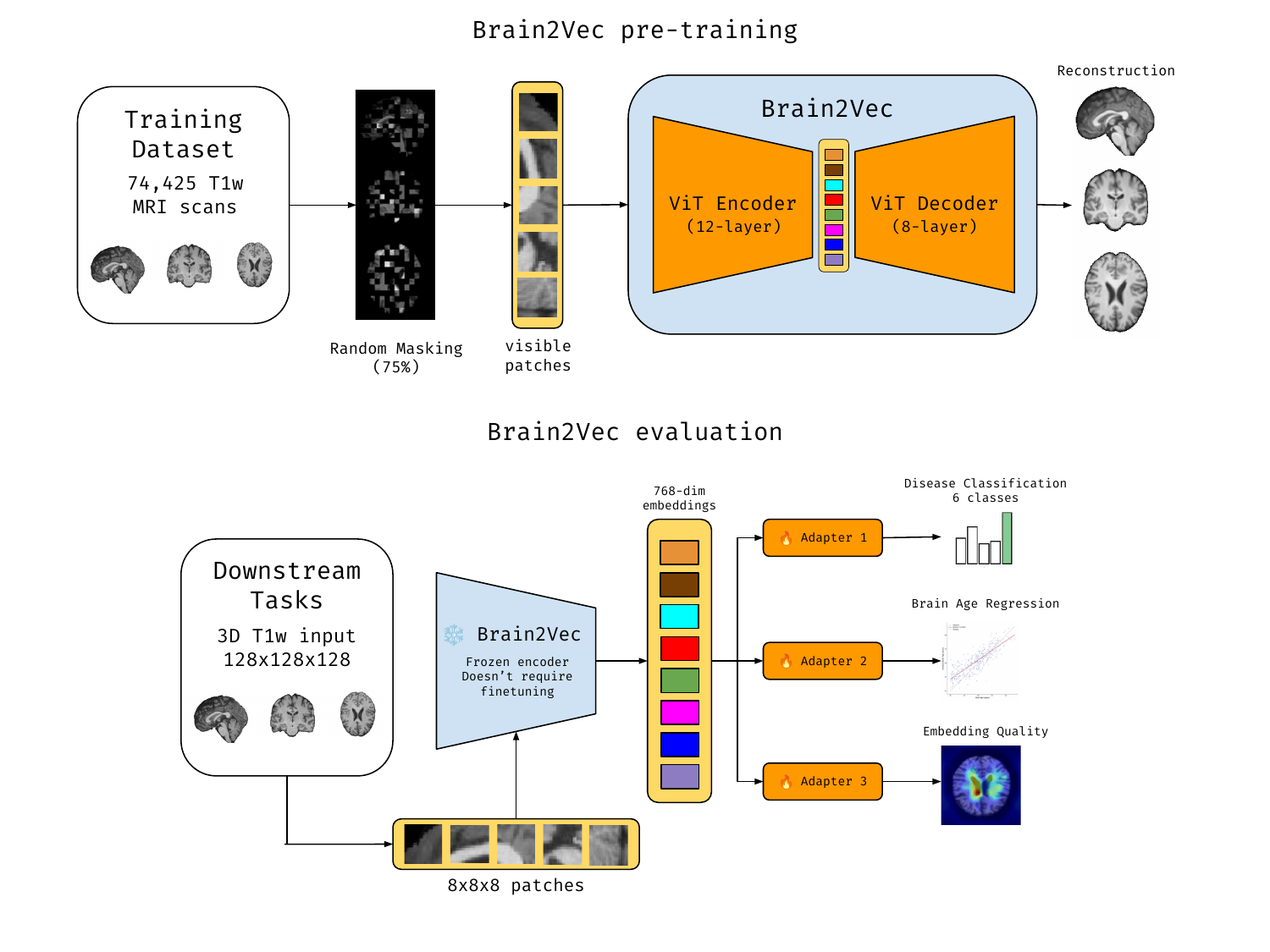}
    \caption{\textbf{Overview of the Brain2Vec pre-training architecture and downstream applications.}
    \textbf{(Pre-training)} Brain2Vec is a 3D Masked Autoencoder (MAE) pre-trained on 74,425 T1-weighted MRI scans from the iSTAGING consortium. Input volumes are divided into $8\times8\times8$ patches, with 75\% randomly masked. A 12-layer Vision Transformer (ViT) encoder processes only the visible patches to generate a 768-dimensional latent representation. Mask tokens are then introduced, and an 8-layer ViT decoder reconstructs the original scan.
    \textbf{(Evaluation)} The pre-trained encoder serves two complementary roles. First, it provides a robust weight initialization for end-to-end fine-tuning on a downstream task (brain age prediction), accelerating convergence and improving accuracy relative to random initialization. Second, its frozen 768-dimensional embeddings serve as fixed features for lightweight downstream models, including a Support Vector Regressor (SVR) for brain age prediction and SVMs for five binary classification tasks: CU vs. AD, elevated BMI, diabetes, smoking status, and hypertension. We additionally assess the quality and interpretability of these frozen embeddings via self-attention visualization, correlation with anatomically-defined Regions of Interest (ROIs), and latent-space manipulation, showing that Brain2Vec encodes spatially localized, neuroanatomically meaningful representations.}
    \label{fig:brain2vec_architecture}
\end{figure}

Brain2Vec is a 3D Masked Autoencoder \citep{he_masked_2022} with an assymetric encoder-decoder design (Fig.~\ref{fig:brain2vec_architecture}).

\paragraph{Feature extraction.} A preprocessed volume $x \in \mathbb{R}^{128\times128\times128}$ is split into $N = 16^3 = 4096$ non-overlapping $8\times8\times8$ patches $\{x_p\}_{p=1}^{N}$, each flattened to a 512-dimensional vector and linearly projected to the encoder width $d=768$. A learnable \texttt{[CLS]} token is prepended to the sequence.

\paragraph{Tokenization.} The encoder is a 12-layer ViT \citep{dosovitskiy_image_2020} with a hidden size of 768, 12 attention heads, and an MLP ratio of 4. A random subset (75\%) of patches are masked and the encoder only processes a small subset of visible patches (25\%), without incorporating mask tokens \citep{devlin_bert_2019}.

\paragraph{Decoder. } The decoder is an 8-layer ViT with an input dimensionality of $d=512$ and 16 attention heads. The encoder's outputs are first projected to the required dimensionality and a shared learnable mask token is inserted at every masked position, followed by positional embeddings. The decoder processes the full sequence of tokens and a linear head predicts the $8\times8\times8= 512$ voxel values of each patch. In all previous work, the decoder is not used for feature extraction but only during pre-training and discarded afterwards. We proposed an approach on using the decoder for interpretability (Sec.~\ref{sec:brain2vec_embedding_analysis}).

\paragraph{Objective.} Let $\hat{x}_p$ denote the predicted voxels of patch $p$. The model is trained to minimize the mean squared error over masked patches only:
\begin{equation}
    \mathcal{L} = \frac{1}{|M|}\sum_{p \in M} \frac{1}{512}\left\lVert \hat{x}_p - x_p \right\rVert_2^2 ,
\end{equation}
where targets $x_p$ are defined in the Z-score-normalized intensity space. Because our goal is a robust latent representation of brain structure rather than recovery of absolute voxel intensities, we do not reconstruct the original intensities.

\paragraph{Embedding extraction. } After pre-training, the decoder is discarded. We freeze the encoder weights and pass all the patches without any masking. The final embedding is obtained from the \texttt{[CLS]} token, though obtaining the final embedding using mean pooling is also a valid option and is used in other recent foundation models like NeuroJEPA.

\subsection{Training}
\label{subsec:brain2vec_training}

\begin{figure}[htbp]
  \centering
  \begin{subfigure}[t]{0.315\linewidth}
    \centering
    {\scriptsize\makebox[0.333\linewidth]{Masked}\makebox[0.333\linewidth]{Recon.}\makebox[0.333\linewidth]{GT}}\\[1pt]
    \includegraphics[width=\linewidth]{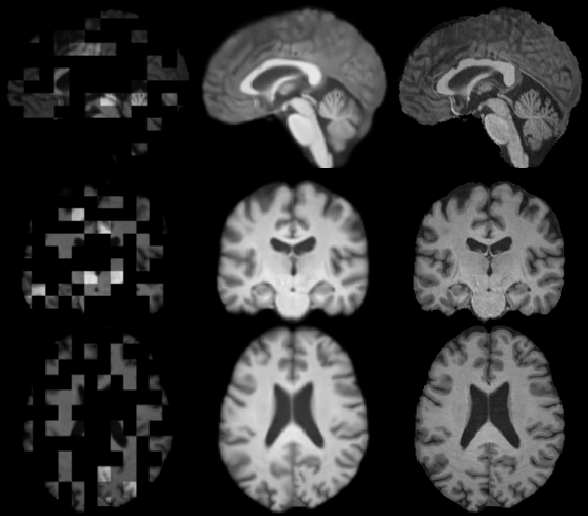}
    \caption{CU}
  \end{subfigure}
  \hfill
  \begin{subfigure}[t]{0.315\linewidth}
    \centering
    {\scriptsize\makebox[0.333\linewidth]{Masked}\makebox[0.333\linewidth]{Recon.}\makebox[0.333\linewidth]{GT}}\\[1pt]
    \includegraphics[width=\linewidth]{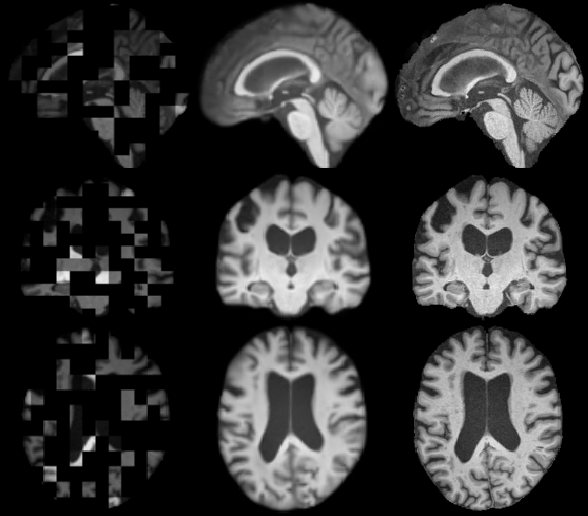}
    \caption{MCI}
  \end{subfigure}
  \hfill
  \begin{subfigure}[t]{0.315\linewidth}
    \centering
    {\scriptsize\makebox[0.333\linewidth]{Masked}\makebox[0.333\linewidth]{Recon.}\makebox[0.333\linewidth]{GT}}\\[1pt]
    \includegraphics[width=\linewidth]{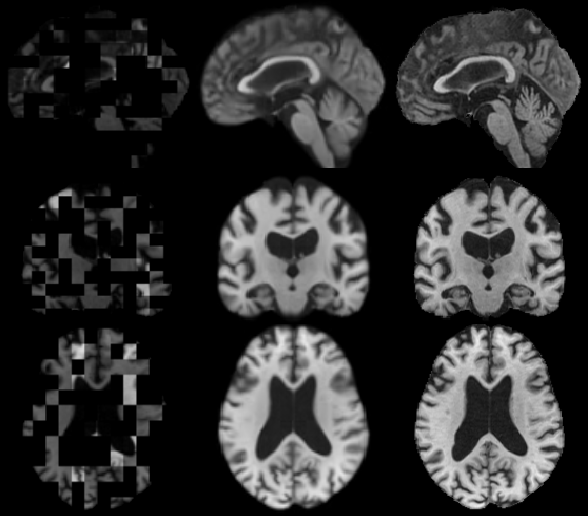}
    \caption{AD}
  \end{subfigure}
  \caption{Brain2Vec reconstruction results (75\% random masking) for a representative subject from each cognitive group: (a) CU (Cognitively Unimpaired), (b) MCI (Mild Cognitive Impairment), and (c) AD (Alzheimer's Disease). Within each panel, columns show the masked input, the Brain2Vec reconstruction (Recon.), and the ground truth (GT); rows show sagittal, coronal, and axial views.}
  \label{fig:reconstruction_results}
\end{figure}

During training, we used $8\times8\times8$ voxel patches with a random masking ratio of 75\%. Fig.~\ref{fig:reconstruction_results} illustrates example scans with random masking applied, alongside their reconstruction generated by Brain2Vec and ground truth.

Data were split into 80\% training, 10\% validation, and 10\% testing sets, stratified by disease. Given the foundation model nature of this work, the training set was intentionally maximized to expose the model to as much diverse data as possible during pre-training, as broader data coverage is critical for learning generalizable representations. Brain2Vec was trained for 500 epochs with an effective batch size of 64 on two A100 GPUs, using the AdamW optimizer \citep{adamw_optimizer} with momentum of $\beta_1 = 0.9, \beta_2 = 0.95$, a base learning rate of $10^{-3}$, and a cosine annealing scheduler without warm restarts \citep{lr_scheduler}. We applied the linear learning rate scaling rule \citep{linear_lr_rule}, $\mathrm{lr} = \mathrm{lr}_{\text{base}} \times B_{\text{eff}} / 32$, where $B_{\text{eff}} = 64$ is the effective batch size, as we trained on two GPUs with a batch size of 32 each.

All scans were skull-stripped using a deep learning–based method and linearly registered to a common atlas space \citep{bashyam_dlmuse_2025, pomponio_harmonization_2020}. Preprocessed scans were resized to $128 \times 128 \times 128$ voxels to reduce computational cost. It is important to note that the objective of our training is not precise reconstruction, but rather learning meaningful representations of the input data. To improve generalization, we applied data augmentation using TorchIO \citep{torchio}, including RandomBlur ($\sigma_i \sim \upsilon(0, 2)$), RandomGamma ($\beta \sim \upsilon(-0.3, 0.3)$), RandomFlip (lateral axis only), RandomNoise (mean=0, std=0), and affine transformations.

\subsection{Evaluation}
The pre-trained Brain2Vec model was evaluated on six downstream tasks including: (i) five classification tasks (CU vs. AD, elevated BMI, diabetes, smoking status and hypertension) and (ii) a brain age prediction task, under two complementary configurations:

\begin{itemize}
  \item \textbf{Encoder fine-tuning:} The pre-trained encoder weights were used as initialization for a Vision Transformer (ViT) applied to a brain age prediction task. This approach preserves the original ViT architecture, allowing the downstream model to leverage the rich structural representations acquired during pre-training.
  
  \item \textbf{Feature extraction:} MRI scans were passed through the frozen pre-trained encoder to extract 768-dimensional embeddings (Brain2Vec features), which subsequently served as fixed input representations for lightweight downstream models using Support Vector Machines (SVMs).
\end{itemize}

We compared fine-tuning from Brain2Vec pretrained weights against training from random initialization, measuring both final accuracy and convergence speed. In the feature extraction setting, we compared downstream predictive performance against SPARE-BA, a model trained using segmentation-based regional volumetric features and using features extracted from NeuroJEPA, a state-of-the-art foundation model, providing a direct assessment of Brain2Vec relative to both traditional and more recent neuroimaging feature pipelines.

Finally, as an external evaluation, Brain2Vec embeddings were assessed on BrainFMBench \citep{Encin2026.05.15.725427} (datasets described in Sec.~\ref{subsec:datasets}). The benchmark probes frozen embeddings with linear and random-forest models on held-out test sets, averaging results over five random seeds, and compares them with other foundation models and with FreeSurfer-derived regional features. We report the linear-probe results. Brain2Vec embeddings were extracted with our own preprocessing (DLICV skull stripping and linear registration to the Brain2Vec template), whereas the other image-based models used the benchmark's turboprep preprocessing.

\section{Results}
\label{sec:brain2vec_results}

\subsection{Fine-tuning}
\label{sec:ba_finetuning}

Brain2Vec encoder weights were used to initialize a ViT-based brain age prediction model (ViT-BA). We fine-tuned the model on our brain age prediction dataset explained in Sec.~\ref{sec:datasets}. In contrast to Brain2Vec model training stage, where maximizing training data sample size was prioritized, a larger proportion of data was used in this validation experiment to ensure robust evaluation of regression performance on the downstream task. Data were partitioned into 60\% for training, 30\% for validation, and 10\% for testing. All training configurations and hyperparameters were kept consistent with those used during the Brain2Vec model training. Results are reported with and without Brain2Vec initialization, enabling a direct assessment of the contribution of transfer learning to downstream regression performance.

\begin{figure}[!tbp]
  \centering

  \begin{subfigure}[b]{\linewidth}
    \includegraphics[width=\linewidth]{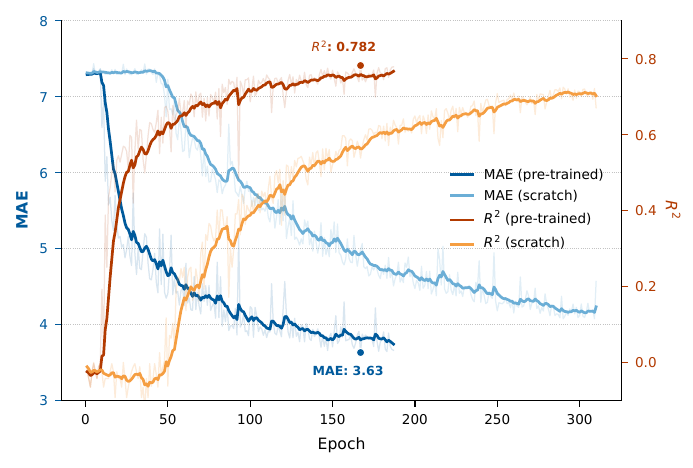}
    \caption{ViT-BA validation metrics per epoch.}
    \label{fig:ba_training_logs}
  \end{subfigure}

  \vspace{0.5em}

  \begin{subfigure}[t]{0.48\linewidth}
      \includegraphics[width=\linewidth]{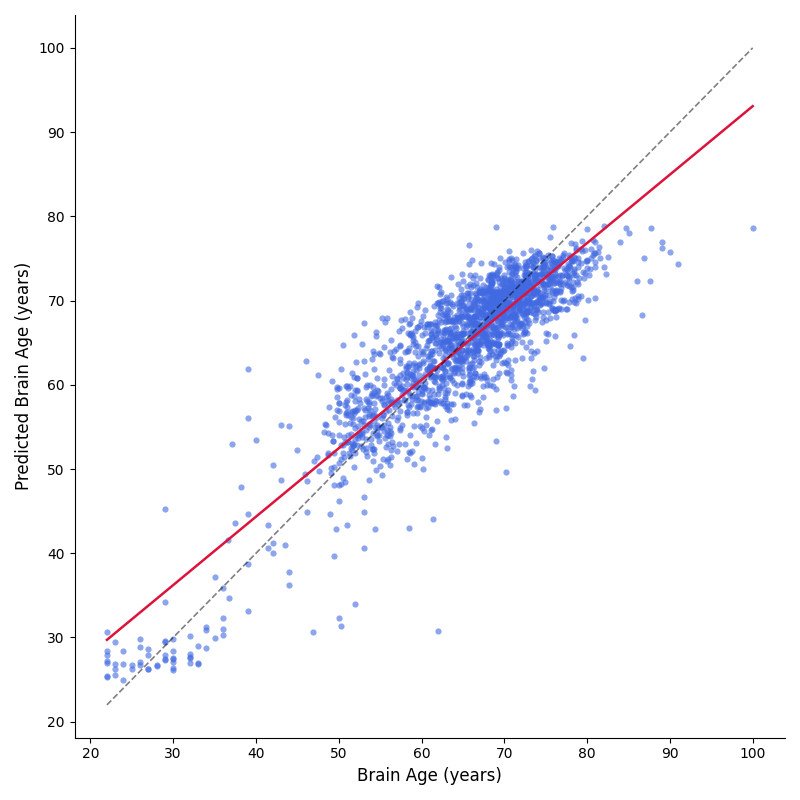}
      \caption{ViT-BA brain age prediction on the held-out set using pre-training.}
      \label{fig:ba_pretrain}
  \end{subfigure}%
  \hfill
  \begin{subfigure}[t]{0.48\linewidth}
      \includegraphics[width=\linewidth]{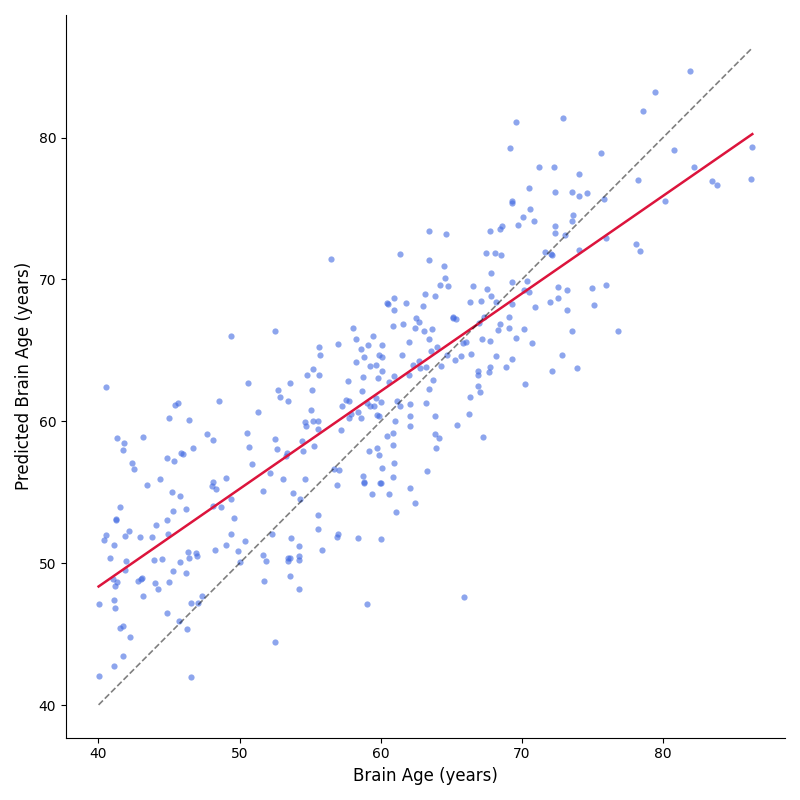}
      \caption{Brain2Vec-SVR brain age prediction on IXI.}
      \label{fig:svr_ba_ixi}
  \end{subfigure}

  \caption{Downstream brain age prediction performance of Brain2Vec.}
  \label{fig:brain_age_prediction}
\end{figure}

As expected from pre-trained initializations, optimization dynamics were markedly improved, with the Brain2Vec-initialized model bypassing the initial plateau within fewer than 10 epochs compared to approximately 50 epochs for the randomly initialized model (Fig.~\ref{fig:ba_training_logs}). This confirms that self-supervised pre-training provides a superior weight initialization that optimizes training efficiency, accelerating convergence.

Brain2Vec initialization also improved brain age prediction accuracy (Table~\ref{tab:brain_age_results}, Fig.~\ref{fig:ba_pretrain}). The pre-trained model outperformed the randomly initialized baseline across all metrics -- Mean Absolute Error (MAE), Mean Squared Error (MSE), Root Mean Squared Error (RMSE), and $R^2$ -- on both the held-out test set and the external IXI dataset~\citep{ixi_dataset}, with the performance gap being more pronounced on the latter. The larger improvement on out-of-sample data suggests that pre-training contributes to better generalization across sites.

\begin{table}[htbp]
  \centering
  \caption{brain age prediction results with and without Brain2Vec pre-training on the held-out test set and the external IXI dataset.}
  \label{tab:brain_age_results}
  \begin{tabular}{llccccc}
    \toprule
    Dataset & Model & Pre-training & MAE $\downarrow$ & MSE $\downarrow$ & RMSE $\downarrow$ & $R^2 \uparrow$ \\
    \midrule
    \multirow{2}{*}{Test} & ViT-BA & No  & 4.07 & 28.50 & 5.34 & 0.73 \\
                          & ViT-BA & Yes & \textbf{3.56} & \textbf{21.99} & \textbf{4.69} & \textbf{0.79} \\
    \midrule
    \multirow{2}{*}{IXI}  & ViT-BA & No  & $6.18$ & $63.93$ & $8.00$ & $0.43$ \\
                          & ViT-BA & Yes & $\bm{5.17}$ & $\bm{44.40}$ & $\bm{6.66}$ & $\bm{0.60}$ \\
    \bottomrule
  \end{tabular}
\end{table}

\subsection{Feature extraction}

\subsubsection{Brain age prediction}
We trained a Support Vector Regressor (SVR) on frozen Brain2Vec encoder embeddings to predict brain age. The rationale is that an effectively pretrained encoder that captures neuroanatomically relevant structure should enable a simple non-parametric model to generalize effectively to age prediction from fixed embeddings alone, without any backbone fine-tuning. We used 5-fold cross-validation for data splitting, with hyperparameters tuned via Optuna using nested 5-fold cross-validation on the training folds, optimizing for $R^2$ score. Splits were randomly sampled.

Our results (Table~\ref{tab:svm_ba_results}, Fig.~\ref{fig:svr_ba_ixi}) demonstrate that the proposed approach achieves performance comparable to fully supervised ViT regressors, despite its simplicity, suggesting that Brain2Vec embeddings capture rich structural information relevant to brain aging. Notably, the SVR model trained on Brain2Vec features outperformed SPARE-BA, a validated and established brain age estimation method trained on volumetric features from anatomical segmentation using a pre-defined atlas \citep{habes_advanced_2016}, and performed comparable to SVR models trained on NeuroJEPA embeddings. This suggests that Brain2Vec features encode structural information that is at least as informative as anatomically defined, handcrafted regional volume features. We also evaluated the utility of brain age gap (predicted age $-$ actual age) to capture disease related or risk-factor effects on multiple disease-specific datasets described in Sec.~\ref{sec:datasets}. The brain age gap from Brain2Vec-SVR model showed significant group differences in AD (Cohen's d = $0.94$, $p < 0.001$) and elevated BMI (Cohen's d = $0.43$, $p < 0.001$), outperforming ViT-BA model on this measure. Results are reported in supplementary Fig.~\ref{fig:ba_gap}.

Validations on the external IXI dataset were performed using the same SVR models trained in the cross-validation folds, without any additional training or fine-tuning. This approach was chosen to directly assess the generalization performance of the learned Brain2Vec feature space without any additional model adaptation, consistent with the frozen feature extraction paradigm. The same inference procedure was applied to SPARE-BA models trained on DLMUSE-derived features and NeuroJEPA-SVR trained on NeuroJEPA embeddings, ensuring a fair comparison between approaches under identical conditions. For the IXI inference, we combined predictions from all 5-fold SVR models into an ensemble, as this approach typically yields more robust and stable predictions by averaging out individual model variances. The results of the IXI inference are included in Table~\ref{tab:svm_ba_results} and Fig.~\ref{fig:svr_ba_ixi}. The SVR model fitted on Brain2Vec embeddings yielded the best results, confirming that Brain2Vec has learned a generalizable representation of brain morphology that serves as a compact yet informative feature space for efficient downstream tasks.

\begin{table}[htbp]
  \centering
  \caption{SVR brain age prediction testing results (mean $\pm$ std across 5 folds) and inference results on IXI.}
  \label{tab:svm_ba_results}
  \resizebox{\linewidth}{!}{%
  \begin{tabular}{lcccc}
    \toprule
    Model & MAE $\downarrow$ & MSE $\downarrow$ & RMSE $\downarrow$ & $R^2 \uparrow$ \\
    \midrule
    SPARE-BA & $4.37 \pm 0.05$ & $33.09 \pm 0.67$ & $5.75 \pm 0.06$ & $0.68 \pm 0.01$ \\
    NeuroJEPA-SVR & $3.44 \pm 0.07$ & $\bm{19.60 \pm 0.79}$ & $\bm{4.43 \pm 0.09}$ & $\bm{0.81 \pm 0.01}$ \\
    Brain2Vec-SVR & $\bm{3.43 \pm 0.02}$ & $19.62 \pm 0.13$ & $\bm{4.43 \pm 0.01}$ & $\bm{0.81 \pm 0.00}$ \\
    \midrule
    \multicolumn{5}{l}{\textit{IXI Inference}} \\
    \midrule
    SPARE-BA & $7.89$ & $93.08$ & $9.65$ & $0.17$ \\
    NeuroJEPA-SVR & $5.19$ & $40.93$ & $6.40$ & $\bm{0.64}$ \\
    Brain2Vec-SVR & $\bm{5.10}$ & $\bm{40.50}$ & $\bm{6.36}$ & $\bm{0.64}$ \\
    \bottomrule
  \end{tabular}
  }
\end{table}

\subsubsection{Disease and risk-factor classification}

Following the same protocol as the brain age prediction experiments, we trained SVM classifiers on our disease classification dataset explained in Sec.~\ref{sec:datasets}. We used 5-fold cross-validation, with hyperparameters tuned via Optuna~\citep{optuna} through nested 5-fold cross-validation on each training set, optimizing for F1 score. Splits were stratified by disease or risk-factor.

Despite the encoder being trained without any disease supervision, the learned representations yielded strong classification performance across all disease groups, indicating that Brain2Vec captures clinically meaningful morphological structure beyond its explicit training objective. Compared to SPARE models trained on DLMUSE-derived features, our approach outperformed across all tasks in terms of accuracy, AUC, F1 score, recall and precision, except for AD classification, where DLMUSE retained a clear advantage. Compared to NeuroJEPA, Brain2Vec performed better on diabetes and smoking classification, while NeuroJEPA performed better on AD and elevated BMI classification, with comparable results on hypertension (Table~\ref{tab:svm_results}).

\begin{table}[H]
  \caption{SVM disease and risk-factor classification testing results (mean $\pm$ std across 5 folds).}
  \centering
  \label{tab:svm_results}
  \resizebox{\linewidth}{!}{%
  \small
  \begin{tabular}{llccccc}
    \toprule
    Task & Method & Accuracy $\uparrow$ & AUC $\uparrow$ & F1 $\uparrow$ & Recall $\uparrow$ & Precision $\uparrow$ \\
    \midrule
    \multirow{3}{*}{CU vs.\ AD}
      & DLMUSE    & $\bm{0.83 \pm 0.02}$ & $\bm{0.90 \pm 0.01}$ & $\bm{0.83 \pm 0.02}$ & $\bm{0.86 \pm 0.03}$ & $0.82 \pm 0.02$ \\
      & NeuroJEPA & $0.80 \pm 0.02$ & $0.88 \pm 0.01$ & $0.80 \pm 0.02$ & $0.77 \pm 0.05$ & $\bm{0.83 \pm 0.02}$ \\
      & Brain2Vec & $0.78 \pm 0.02$ & $0.86 \pm 0.02$ & $0.78 \pm 0.02$ & $0.77 \pm 0.02$ & $0.79 \pm 0.04$ \\
    \midrule
    \multirow{3}{*}{CU vs.\ Elevated BMI}
      & DLMUSE    & $0.61 \pm 0.02$ & $0.65 \pm 0.04$ & $0.61 \pm 0.02$ & $0.63 \pm 0.02$ & $0.60 \pm 0.03$ \\
      & NeuroJEPA & $\bm{0.79 \pm 0.02}$ & $\bm{0.87 \pm 0.01}$ & $\bm{0.79 \pm 0.02}$ & $\bm{0.79 \pm 0.02}$ & $\bm{0.79 \pm 0.02}$ \\
      & Brain2Vec & $0.78 \pm 0.01$ & $0.85 \pm 0.01$ & $0.78 \pm 0.01$  & $0.76 \pm 0.01$ & $0.78 \pm 0.01$ \\
    \midrule
    \multirow{3}{*}{CU vs.\ Diabetes}
      & DLMUSE    & $0.59 \pm 0.01$ & $0.64 \pm 0.01$ & $0.59 \pm 0.01$ & $0.61 \pm 0.02$ & $0.59 \pm 0.01$ \\
      & NeuroJEPA & $0.61 \pm 0.06$ & $0.65 \pm 0.08$ & $0.61 \pm 0.06$ & $\bm{0.64 \pm 0.03}$ & $0.61 \pm 0.06$ \\
      & Brain2Vec & $\bm{0.64 \pm 0.01}$ & $\bm{0.70 \pm 0.01}$ & $\bm{0.64 \pm 0.01}$ & $0.62 \pm 0.01$ & $\bm{0.65 \pm 0.01}$\\
    \midrule
    \multirow{3}{*}{CU vs.\ Smoking}
      & DLMUSE    & $0.51 \pm 0.02$ & $0.49 \pm 0.01$ & $0.50 \pm 0.02$ & $0.63 \pm 0.07$ & $0.51 \pm 0.02$ \\
      & NeuroJEPA & $0.54 \pm 0.04$ & $0.55 \pm 0.05$ & $0.53 \pm 0.04$ & $0.57 \pm 0.05$ & $0.53 \pm 0.04$ \\
      & Brain2Vec & $\bm{0.62 \pm 0.01}$ & $\bm{0.66 \pm 0.02}$ & $\bm{0.62 \pm 0.02}$ & $\bm{0.65 \pm 0.02}$ & $\bm{0.62 \pm 0.02}$ \\
    \midrule
    \multirow{3}{*}{CU vs.\ Hypertension}
      & DLMUSE    & $0.52 \pm 0.03$ & $0.50 \pm 0.00$ & $0.52 \pm 0.03$ & $0.61 \pm 0.04$ & $0.52 \pm 0.03$ \\
      & NeuroJEPA & $\bm{0.61 \pm 0.01}$ & $\bm{0.65 \pm 0.01}$ & $0.61 \pm 0.01$ & $0.64 \pm 0.02$ & $\bm{0.61 \pm 0.01}$ \\
      & Brain2Vec & $0.60 \pm 0.01$ & $0.64 \pm 0.01$ & $\bm{0.62 \pm 0.01}$ & $\bm{0.65 \pm 0.03}$ & $0.59 \pm 0.02$ \\
    \bottomrule
  \end{tabular}
  }
\end{table}

We further assessed generalization through external validation on the MIRIAD dataset~\citep{malone2013miriad}, comprising baseline T1 MRI scans from 46 mild-to-moderate Alzheimer's subjects and 23 cognitively normal controls. The DLMUSE-based features achieved the best overall performance. Brain2Vec reached an accuracy comparable to NeuroJEPA (0.84 vs. 0.83) with substantially higher recall (0.93 vs. 0.76) but lower precision (0.85 vs. 0.99), while NeuroJEPA achieved a higher AUC (0.97 vs. 0.94). Therefore, these differences in recall and precision largely reflect different operating points under potential domain shift rather than differences in discriminative ability: NeuroJEPA ranked the cases well but predicted AD conservatively at the default threshold, whereas Brain2Vec favored sensitivity. Overall, Brain2Vec generalized reasonably well to this independent cohort, although threshold calibration on new cohorts may be needed for all embedding-based models.

\begin{table}[htbp]
  \centering
  \caption{CU vs AD inference on MIRIAD (mean $\pm$ std across 5 folds).}
  \label{tab:miriad_cn_ad}
  \resizebox{\linewidth}{!}{%
  \begin{tabular}{lccccc}
    \toprule
    Model & Accuracy $\uparrow$ & AUC $\uparrow$ & F1 $\uparrow$ & Recall $\uparrow$ & Precision $\uparrow$ \\
    \midrule
    DLMUSE & $\bm{0.94 \pm 0.01}$ & $\bm{0.98 \pm 0.01}$ & $\bm{0.93 \pm 0.01}$ & $\bm{0.93 \pm 0.01}$ & $0.97 \pm 0.01$ \\
    NeuroJEPA & $0.83 \pm 0.04$ & $0.97 \pm 0.00$ & $0.82 \pm 0.05$ & $0.76 \pm 0.06$ & $\bm{0.99 \pm 0.01}$ \\
      Brain2Vec & $0.84 \pm 0.02$ & $0.94 \pm 0.01$ & $0.81 \pm 0.03$ & $\bm{0.93 \pm 0.02}$ & $0.85 \pm 0.03$ \\
    \bottomrule
  \end{tabular}
  }
\end{table}

We report per-site classification results in supplementary Table~\ref{tab:per_site_results} for the three sites with the largest sample sizes. Results are broadly consistent with the overall findings in Table~\ref{tab:svm_results}, while UKBiobank data, with the largest sample size obtained higher accuracy, and sites with younger subjects that fall outside the Brain2Vec model's training distribution, such as CARDIA and HCP, obtained lower accuracy. This highlights the importance of training data coverage for foundation models, as well as the potential need for domain adaptation techniques to optimize performance on out-of-distribution cohorts. 

Supplementary Fig.~\ref{fig:pca_batch_effects} presents PCA visualizations of Brain2Vec embeddings for each classification cohort, colored by study site. The embeddings exhibit site-related structure, indicating that Brain2Vec also captures study-level variation in the input data.

\subsection{BrainFMBench Results}
\label{subsec:brainfmbench_results}

\begin{figure*}[htbp]
    \centering
    \includegraphics[width=\textwidth,height=0.85\textheight,keepaspectratio]{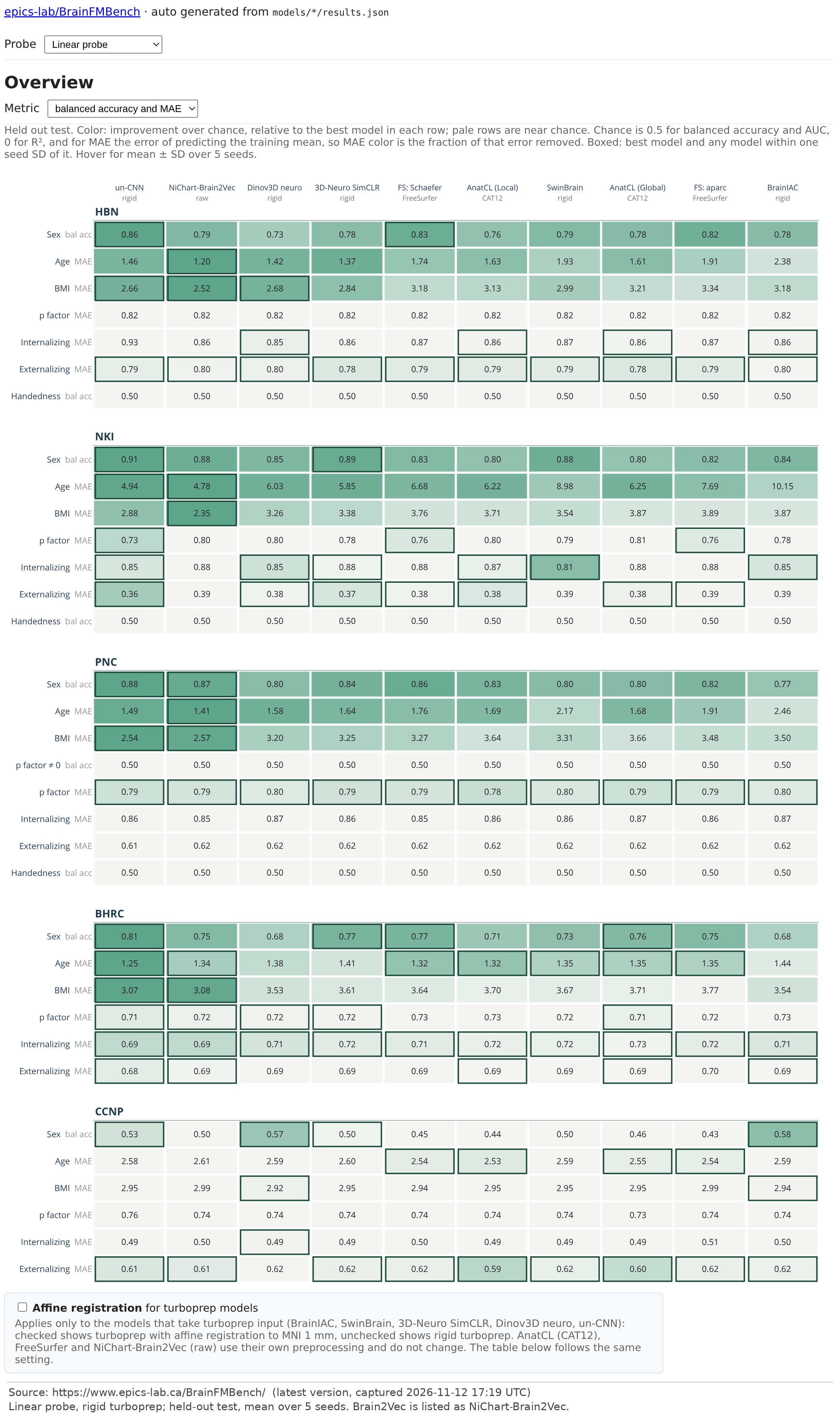}
    \caption{\textbf{BrainFMBench leaderboard (linear probe).} Held-out test results, averaged over five seeds, for each cohort (rows grouped by study) and model (columns); Brain2Vec is listed as NiChart-Brain2Vec. Cell color indicates improvement over chance relative to the best model in each row (pale rows are near chance), and boxes mark the best model and any model within one seed standard deviation of it. Classification tasks report balanced accuracy and regression tasks MAE. Image generated from the benchmark's leaderboard, \url{https://www.epics-lab.ca/BrainFMBench/} (v2.2.0, linear probe), captured 12 November 2026, 17:19 UTC.}
    \label{fig:brainfmbench_leaderboard}
\end{figure*}

Fig.~\ref{fig:brainfmbench_leaderboard} shows the BrainFMBench linear-probe results for all models and tasks. For the sex, age, and BMI prediction tasks across the five cohorts, Brain2Vec ranked first among the six pretrained foundation models (counting the Global and Local AnatCL variants as one method) on 9 of 15 tasks and first or second on 11 of 15; results were similar with the random-forest probe (first on 8 of 15). It obtained the lowest error of all entries for age prediction in HBN, NKI, and PNC and for BMI prediction in HBN and NKI. The benchmark maintainers' baseline, un-CNN, which uses substantially higher-dimensional features (9,664 versus 768), obtained the best sex classification results and several of the best results in BHRC. Psychopathology scores and handedness were near chance level for all models, and no model predicted age or BMI in CCNP, so these tasks do not differentiate between models.

\subsection{Interpretability of Brain2Vec's Learned Representations}
\label{sec:brain2vec_embedding_analysis}

In medical imaging, there is a growing need for interpretable models, as they enable researchers and clinicians to understand how a model makes predictions and which features are important for each task \citep{vanveldhuizen2025foundationmodelsmedicalimaging}. We assessed Brain2Vec's interpretability through three complementary strategies: (i) visualizing self-attention maps to examine which spatial regions the model attends to, (ii) correlating latent embeddings with anatomically-defined regions of interest to identify which brain structures drive specific features, and (iii) manipulating latent embeddings and inspecting their effect on reconstructed anatomy to test whether individual features encode monotonic, region-specific changes.

\begin{figure}[!tbp]
\centering
\includegraphics[width=\textwidth]{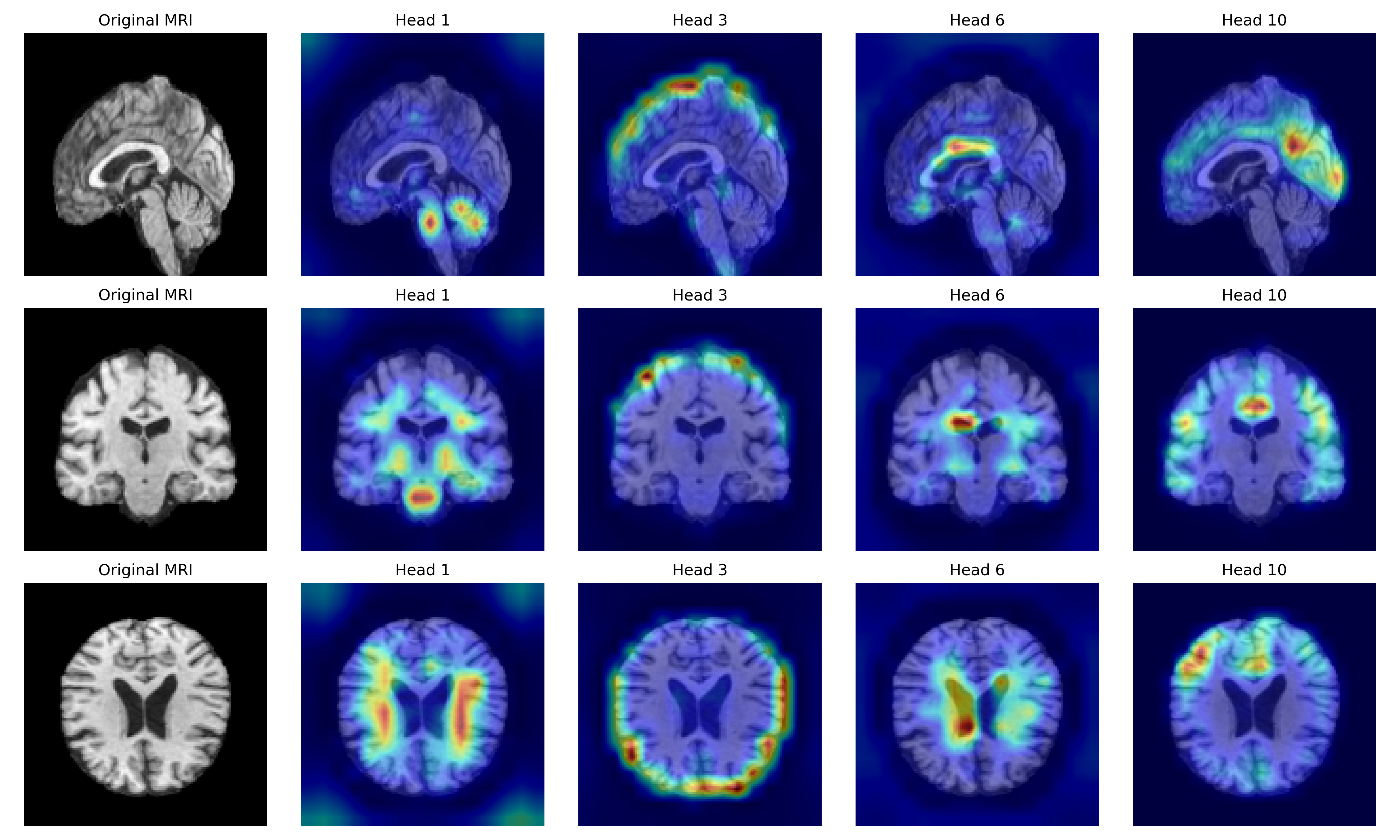}
\caption{Self-attention map of a participant with Alzheimer's Disease derived using the Vision Transformer (ViT) encoder of Brain2Vec. The attention map is computed using the \texttt{[CLS]} token on the heads of the encoder's last ViT layer. Additional examples for all the heads are presented in supplementary Section~\ref{sec:self_attention}.}
\label{fig:self_attention}
\end{figure}

\paragraph{Attention maps.} We visualized what the model considers important by plotting its self-attention maps (Fig.~\ref{fig:self_attention}). The resulting maps show that different attention heads learn distinct spatial patterns rather than attending uniformly across the entire brain, consistent with findings in other models \citep{caron2021emergingpropertiesselfsupervisedvision}. As shown in the attention visualizations, different heads consistently emphasize different anatomical regions across the axial, coronal, and sagittal views. Some heads exhibit relatively localized attention around specific brain structures, whereas others capture more distributed patterns across larger portions of the brain. This diversity suggests that the multi-head self-attention mechanism enables Brain2Vec to encode complementary spatial information, with individual heads potentially capturing different anatomical or contextual characteristics of the input MRI. An important exception is the behavior of heads 0 and 9: across views and samples, these heads frequently assign substantial attention to regions outside the brain, particularly along the image boundaries and background (see Section~\ref{sec:self_attention}). This indicates that they are not primarily encoding meaningful brain anatomy and may instead capture image-level or acquisition-related characteristics.

\begin{figure}[htbp]
    \begin{subfigure}[t]{0.48\linewidth}
      \includegraphics[width=\linewidth]{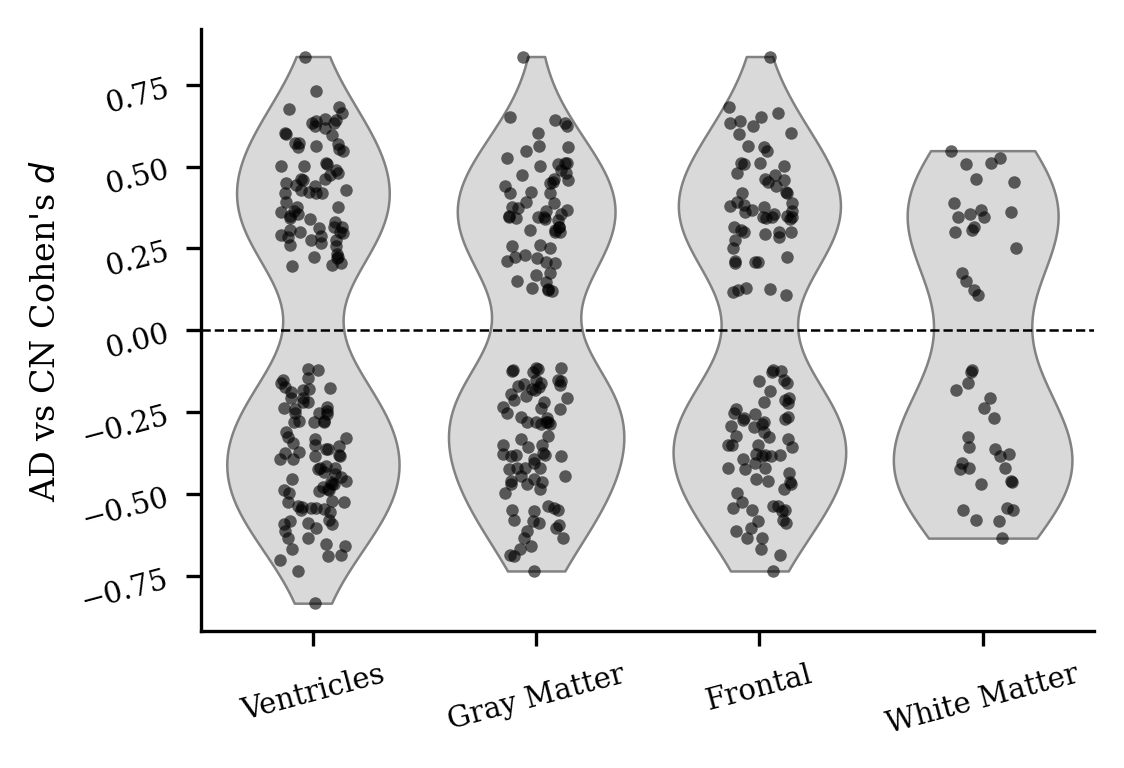}
      \caption{Effect sizes of anatomically matched Brain2Vec dimensions by brain region (Pearson $|r| \geq 0.4$).}
      \label{fig:ad_cn_effect}
    \end{subfigure}
    \hfill
    \begin{subfigure}[t]{0.48\linewidth}
      \includegraphics[width=\linewidth]{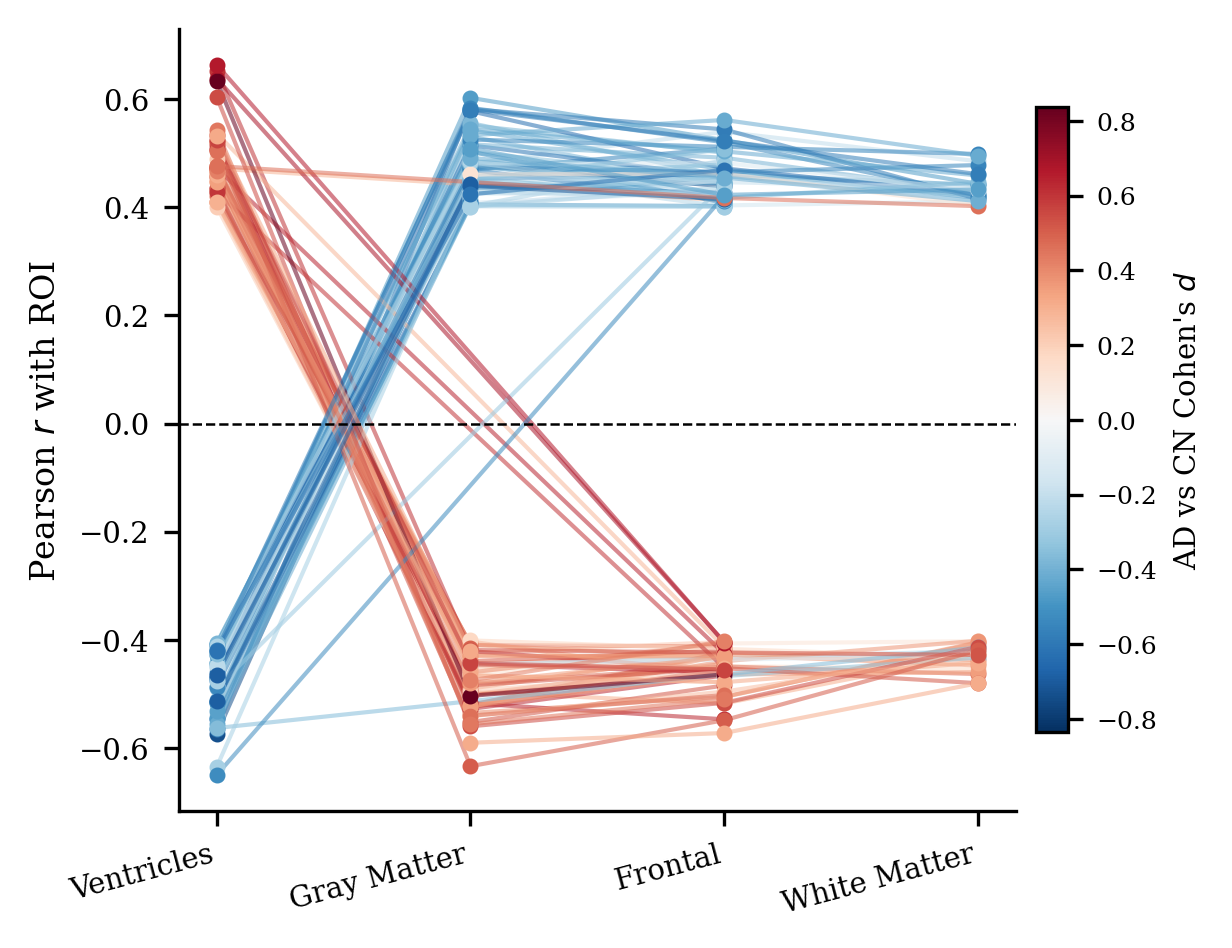}
      \caption{Cross-region correlation profiles of Brain2Vec dimensions.}
      \label{fig:region_profile}
    \end{subfigure}
    \caption{Correlation analysis of Brain2Vec latent embeddings with brain regions.}
    \label{fig:latent_analysis}
\end{figure}

\paragraph{Correlation with anatomically defined regions.} The latent embeddings of foundation models trained on domain-specific data can be interpreted by quantifying their correlation with anatomically defined regions of interest, particularly those known to be affected by disease. To this end, we computed the Pearson correlation between each individual Brain2Vec latent dimension and each DLMUSE regional volume, together with the mean difference in each latent dimension between CU and AD subjects, in order to identify which dimensions are most strongly associated with which anatomical region. As shown in Fig.~\ref{fig:latent_analysis}, several latent dimensions exhibit strong correlations with specific brain regions, including the ventricles, gray matter, frontal regions, and white matter. Consistent with known neurodegenerative patterns in AD, we observe an inverse relationship between ventricular volume and gray matter, frontal, and white matter volumes.

\paragraph{Latent manipulation.} Leveraging the generative nature of the model, we masked brain regions and manipulated latent embeddings to test whether this produces a monotonic effect on the reconstruction. We performed this analysis for three regions: (i) ventricles (Fig.~\ref{fig:ventricles_brain2vec_effect}), (ii) frontal regions (Fig.~\ref{fig:frontal_brain2vec_effect}), and (iii) gray matter (GM) regions (Fig.~\ref{fig:gm_wm_brain2vec_effect}). For masking, a patch was masked if it contained at least one voxel within the region of interest, and otherwise remained visible to the encoder. Detailed masking and reconstruction examples are shown in Figures~\ref{fig:masking_ventricles}, \ref{fig:masking_frontal}, and \ref{fig:masking_gm}. We manually selected a subset $S$ of 22 latent features to manipulate (Supplementary Table~\ref{tab:brain2vec_important_features}). We progressively added the mean difference $\Delta = Z_{AD}[i] - Z_{CU}[i]$, where $Z$ denotes the Brain2Vec latent embeddings, with magnitude $\alpha \in \{0, \ldots, 2.5\}$, giving $\hat{Z}[i] = Z[i] + \alpha \cdot \Delta, \forall i \in S$, where $\hat{Z}$ is the manipulated embedding. After reconstruction, we computed each region's volume using DLMUSE. Across all three regions, we observed the expected monotonic change in regional volume as $\alpha$ increases, confirming that the manipulated latent dimensions encode a consistent, region-specific anatomical effect (Section~\ref{sec:embedding_manipulation}).

\subsection{Inference Speed and Scalability}

Inference efficiency is a critical but often underreported aspect of deep learning evaluation, particularly in medical imaging, where 3D volumetric data introduces substantial computational overhead. For Brain2Vec, which processes full 3D MRI scans to produce subject-level embeddings, efficient inference is especially important given the large and diverse cohorts it targets.

We compared Brain2Vec's inference time against DLMUSE, which uses an nnU-Net~\citep{isensee2018nnunetselfadaptingframeworkunetbased} to segment the brain and compute ROI volumes from the full 3D input, and against NeuroJEPA, which uses a similar ViT encoder augmented with a Mixture of Experts (MoE) layer~\citep{shazeer2017outrageouslylargeneuralnetworks} at inference time. Benchmarks were conducted on the IXI dataset ($581$ scans) using an A40 GPU, as well as single-subject inference on CPU. 
On GPU, Brain2Vec processed the full IXI cohort in $30$ seconds ($\approx 0.05$ seconds per scan), an $\approx 87\times$ speedup over DLMUSE and an $\approx 7\times$ speedup over NeuroJEPA. On CPU, Brain2Vec remained $\approx 8\times$ faster than DLMUSE, whereas NeuroJEPA was faster than Brain2Vec ($6.98$ vs.\ $45.73$ seconds). This is likely because NeuroJEPA's sparse MoE layer activates only a subset of its parameters for each input, which reduces sequential computation on CPU. Brain2Vec's dense ViT encoder, by contrast, benefits more from GPU parallelism. Overall, Brain2Vec offers substantially lower computational cost in GPU-based, cohort-scale settings while maintaining strong predictive performance (Table~\ref{tab:brain2vec_dlmuse_efficiency}). 

\begin{table}[htbp]
  \centering
  \caption{Brain2Vec inference performance.}
  \label{tab:brain2vec_dlmuse_efficiency}
  \begin{tabular}{lcccc}
    \toprule
    Model & FLOPs $\downarrow$ & \#Params & GPU (A40) & CPU \\
    \midrule
    DLMUSE    & 965.3B & 16.6M & 43.41 (min.) - $n = 581$ & 6.39 (min.) - $n = 1$ \\
    NeuroJEPA & \textbf{172.9B} & 122.0M & 3.62 (min.) - $n = 581$ & \textbf{0.12} (min.) - $n = 1$ \\
    Brain2Vec & 177.3B & 88.6M & \textbf{0.50} (min.) - $n = 581$ & 0.76 (min.) - $n = 1$ \\
    \bottomrule
  \end{tabular}
\end{table}

\section{Discussion}
\label{sec:discussion}

In this work, we introduced Brain2Vec, a self-supervised foundation model trained on $74,425$ T1-weighted brain MRI scans. We evaluated Brain2Vec across multiple settings: fine-tuning for brain age prediction, feature extraction for disease classification, external validation on an independent cohort and on the independent BrainFMBench benchmark, interpretability analysis, and inference efficiency. We demonstrated that Brain2Vec serves as a versatile dual-purpose framework: (1) accelerating convergence and improving predictive accuracy when fine-tuned as a 3D ViT backbone, and (2) offering a rich, 768-dimensional frozen feature space that is competitive with or superior to both anatomically-derived volumetric features and larger foundation models across a range of disease classification tasks. On BrainFMBench, Brain2Vec was also among the strongest of the six pretrained foundation models for predicting age, sex, and BMI. Beyond predictive performance, we showed that Brain2Vec's learned representations are anatomically meaningful and interpretable, with generative manipulation of its latent dimensions revealing biologically interpretable, continuous neurodegenerative phenotypes in reconstructed brain space.

Our results reveal a consistent trade-off between anatomical segmentations and self-supervised continuous embeddings. While traditional pipelines like DLMUSE excel at detecting localized structural changes such as volumetric atrophy in Alzheimer's Disease, they are constrained by fixed anatomical boundaries. Conversely, Brain2Vec representations outperformed DLMUSE on harder classification tasks (e.g., diabetes, smoking, elevated BMI). This suggests that Brain2Vec may be learning representations beyond those captured by standard volumetric features derived from anatomical segmentation.

One notable finding of this study is that Brain2Vec-SVR models achieve competitive predictive performance relative to NeuroJEPA, a state-of-the-art foundation model, across several downstream tasks, despite Brain2Vec being pre-trained on $\approx 9.2\times$ fewer scans ($74,425$ vs $683,860$). We attribute this sample efficiency to the inductive bias of our training objective within the structural MRI domain. At their core, Joint Embedding Predictive Architectures (JEPA) like NeuroJEPA, use a context input $x$ to estimate the embedded state of a target input $y$. This process relies on a predictor conditioned on an extra variable $z$ to account for variability during feature-space prediction. However, in standardized neuroimaging datasets where scans are skull-stripped, atlas-registered, and anatomically aligned, latent representation matching presents a relatively low-variance objective. It is well documented that representation collapse is a concern with JEPA architectures~\citep{assran2023selfsupervisedlearningimagesjointembedding}. In contrast, representation collapse is not a concern with reconstruction-based methods. Generative architectures mask a high percentage of the input and use an encoder to produce latent embeddings $z$ of the masked input $y$ and using a decoder, learn to reconstruct the original (unmasked) input $x$. This forces the encoder to learn high-fidelity structural priors and inter-regional anatomical dependencies across the entire brain volume. This dense supervisory signal explains both the model's high sample efficiency and the strong, explicit alignment observed between Brain2Vec latent dimensions and regional brain volumes (Section~\ref{sec:brain2vec_embedding_analysis}).

A common critique of deep feature extractors in neuroimaging is their limited interpretability relative to anatomically defined ROI volumes. Through latent manipulation experiments (Section~\ref{sec:brain2vec_embedding_analysis}), we provided evidence that Brain2Vec's latent dimensions can be linked to anatomically meaningful, region-specific effects, offering a step toward addressing this limitation. Progressively shifting latent dimensions along disease-associated vectors produced monotonic, anatomically plausible changes in reconstructed scans, specifically capturing ventricular expansion, and gray and white matter volume decrease. This generative steering provides evidence that the learned latent space encodes coherent, continuous radiological phenotypes rather than uninterpretable noise.

Despite being trained on a large and diverse dataset, Brain2Vec has two notable limitations. First, UK Biobank dataset dominates the training data ($\approx$ 55\%), which may bias learned representations toward its specific acquisition protocol and reduce embedding quality for scans acquired under substantially different conditions. Second, the training cohort is skewed toward older adults (mean age $65.0$ years; fewer than $5\%$ of scans from studies with mean age below $45$),  which may limit the model's representational accuracy for younger populations. The BrainFMBench cohorts, which are largely pediatric and young-adult, therefore also test generalization beyond the pre-training age range; Brain2Vec performed strongly on most of them, whereas no model, including Brain2Vec, performed above chance on the CCNP cohort. Users applying Brain2Vec outside its training distribution should take these factors into account.

Brain2Vec code and trained models are publicly available at \url{https://github.com/CBICA/Brain2Vec}. Embeddings can be extracted locally via the installable package or through the NiChart cloud platform (\url{https://cloud.neuroimagingchart.com}), which requires no local installation or specialized infrastructure.

\section*{Declaration of the use of AI}
AI tools were used to check spelling and grammar.

\section*{Data and Code Availability}
The codebase and pre-trained weights in safetensors format for both feature extraction and downstream predictive tasks are released as open-source packages on GitHub \url{https://github.com/CBICA/Brain2Vec} and Hugging Face \url{https://huggingface.co/cbica/Brain2Vec}. All tools are also integrated into the NiChart platform \url{https://cloud.neuroimagingchart.com}, enabling users to upload raw T1-weighted MRI scans via the cloud and obtain Brain2Vec embeddings (e.g., for custom model development) as well as a set of biomarkers for disease prediction and brain age estimation.

The iSTAGING dataset used to pre-train Brain2Vec is privately curated. The IXI dataset is publicly available at \url{http://brain-development.org/}. The MIRIAD dataset is publicly available at \url{https://www.ucl.ac.uk/brain-sciences/ion/research/research-centres/dementia-research-centre/research-clinical-trials/minimal-interval-resonance-imaging-alzheimers-disease-miriad}. NKI, HBN, PNC, BHRC and CNC datasets are publicly available at \url{https://reprobrainchart.github.io/}.

\section*{Author Contributions}
Spiros Maggioros: Conceptualization; Methodology; Software; Validation; Formal analysis; Investigation; Visualization; Writing - original draft; and Writing - review \& editing. Guray Erus: Conceptualization; Writing - original draft; and Writing - review \& editing. Gareth Harman: Data Curation. George Aidinis: Conceptualization; and Writing - review \& editing. Alexander Getka: Software. Aristeidis Sotiras: Writing - review \& editing. Christos Davatzikos: Supervision; Funding acquisition; Resources; and Writing - review \& editing.  

\section*{Funding}
This work was supported by the National Institutes of Health grants RF1 AG054409 and U24 NS130411.

\section*{Declaration of Competing Interests}
The authors declare no competing interests.

\printbibliography

\appendix

\renewcommand{\thesection}{\suppletter}
\renewcommand{\thesubsection}{\suppletter.\arabic{subsection}}
\renewcommand{\thetable}{\suppletter\arabic{table}}
\renewcommand{\thefigure}{\suppletter\arabic{figure}}
\renewcommand{\theHtable}{\suppletter\arabic{table}}
\renewcommand{\theHfigure}{\suppletter\arabic{figure}}
\setcounter{table}{0}
\setcounter{figure}{0}

\section{Supplementary material}

\subsection{Characteristics of the datasets used for downstream prediction tasks}
\begin{table}[H]
\centering
\small
\caption{Characteristics of the training and validation datasets for brain age prediction.}
\label{tab:ba_data_overview}
\resizebox{\textwidth}{!}{%
\begin{tabular}{l r c r @{\hspace{1.2cm}} l r c r}
\toprule
\multicolumn{4}{c}{\textit{Training Set}} &
\multicolumn{4}{c}{\textit{Validation Set}} \\
\cmidrule(lr){1-4}\cmidrule(lr){5-8}
\textbf{Study} & \textbf{n} & \textbf{Mean Age $\pm$ Std} & \textbf{(\%) F} &
\textbf{Study} & \textbf{n} & \textbf{Mean Age $\pm$ Std} & \textbf{(\%) F} \\
\midrule
UKBIOBANK  & 10,240 & $66.23 \pm 7.14$  & 45.39 & UKBIOBANK  & 5,141 & $66.25 \pm 7.14$  & 45.75 \\
HCP        & 465    & $42.03 \pm 18.89$ & 54.19 & HCP        & 204   & $41.83 \pm 19.46$ & 58.33 \\
AIBL       & 183    & $72.92 \pm 5.92$  & 55.74 & AIBL       & 90    & $73.03 \pm 5.62$  & 60.00 \\
BLSA       & 158    & $56.71 \pm 17.72$ & 65.19 & BLSA       & 76    & $54.95 \pm 17.63$ & 63.16 \\
SHIP       & 149    & $40.24 \pm 13.37$ & 42.95 & SHIP       & 88    & $38.95 \pm 12.38$ & 34.09 \\
BIOCARD    & 117    & $60.39 \pm 9.83$  & 58.12 & BIOCARD    & 53    & $56.17 \pm 8.24$  & 67.92 \\
PreventAD  & 116    & $62.20 \pm 4.34$  & 68.97 & PreventAD  & 66    & $62.61 \pm 3.92$  & 66.67 \\
ADNI       & 115    & $70.84 \pm 4.05$  & 73.91 & ADNI       & 51    & $72.23 \pm 5.08$  & 64.71 \\
MESA       & 129    & $80.36 \pm 7.45$  & 57.36 & MESA       & 69    & $78.03 \pm 7.87$  & 40.58 \\
CARDIA     & 59     & $51.24 \pm 3.71$  & 44.07 & CARDIA     & 34    & $51.74 \pm 3.08$  & 44.12 \\
OASIS      & 45     & $60.32 \pm 8.56$  & 51.11 & OASIS      & 19    & $62.26 \pm 7.10$  & 73.68 \\
WHIMS      & 14     & $70.77 \pm 4.16$  & 100.00 & WHIMS      & 6     & $69.17 \pm 2.41$  & 100.00 \\
HANDLS     & 8      & $53.75 \pm 8.12$  & 37.50 & HANDLS     & 5     & $53.24 \pm 11.08$ & 0.00 \\
\midrule
\textbf{Total} & \textbf{11,798 (60\%)} & $\bm{64.93 \pm 10.18}$ & \textbf{46.97} &
\textbf{Total} & \textbf{5,902 (30\%)} & $\bm{64.91 \pm 10.15}$ & \textbf{47.09} \\
\bottomrule
\end{tabular}%
}
\end{table}    

\begin{table}[H]
\centering
\small
\caption{Characteristics of the training and validation datasets for CU vs. AD classification.}
\label{tab:dx_cn_ad}
\resizebox{\textwidth}{!}{%
\begin{tabular}{l r c r @{\hspace{1.2cm}} l r c r}
\toprule
\multicolumn{4}{c}{\textit{CU}} &
\multicolumn{4}{c}{\textit{AD}} \\
\cmidrule(lr){1-4}\cmidrule(lr){5-8}
\textbf{Study} & \textbf{n} & \textbf{Mean Age $\pm$ Std} & \textbf{(\%) F} &
\textbf{Study} & \textbf{n} & \textbf{Mean Age $\pm$ Std} & \textbf{(\%) F} \\
\midrule
ADNI      & 419 & $74.68 \pm 7.12$ & 45.11 & ADNI      & 418 & $74.83 \pm 7.72$ & 44.26 \\
OASIS     & 313 & $75.04 \pm 7.33$ & 52.40 & OASIS     & 313 & $75.49 \pm 7.49$ & 49.52 \\
AIBL      & 92  & $74.16 \pm 7.72$ & 59.78 & AIBL      & 92  & $74.23 \pm 7.92$ & 59.78 \\
BLSA      & 9   & $81.33 \pm 8.76$ & 33.33 & BLSA      & 9   & $81.44 \pm 8.91$ & 33.33 \\
PENN      & 7   & $70.01 \pm 7.45$ & 42.86 & PENN      & 7   & $70.81 \pm 6.96$ & 71.43 \\
UKBIOBANK & 4   & $60.87 \pm 4.35$ & 25.00 & UKBIOBANK & 4   & $60.96 \pm 4.43$ & 25.00 \\
BIOCARD   & 1   & $70.37 \pm 0.00$ & 0.00  & BIOCARD   & 1   & $70.26 \pm 0.00$ & 0.00  \\
\midrule
\textbf{Total} & \textbf{845} & $\bm{74.72 \pm 7.39}$ & \textbf{49.11} &
\textbf{Total} & \textbf{844} & $\bm{74.97 \pm 7.76}$ & \textbf{47.87} \\
\bottomrule
\end{tabular}%
}
\end{table}
 
\begin{table}[H]
\centering
\small
\caption{Characteristics of the training and validation datasets for CU vs. Diabetes classification.}
\label{tab:dx_cn_diabetes}
\resizebox{\textwidth}{!}{%
\begin{tabular}{l r c r @{\hspace{1.2cm}} l r c r}
\toprule
\multicolumn{4}{c}{\textit{CU}} &
\multicolumn{4}{c}{\textit{Diabetes}} \\
\cmidrule(lr){1-4}\cmidrule(lr){5-8}
\textbf{Study} & \textbf{n} & \textbf{Mean Age $\pm$ Std} & \textbf{(\%) F} &
\textbf{Study} & \textbf{n} & \textbf{Mean Age $\pm$ Std} & \textbf{(\%) F} \\
\midrule
UKBIOBANK & 2,051 & $66.12 \pm 7.16$  & 36.03  & UKBIOBANK & 2,051 & $66.14 \pm 7.19$  & 36.03  \\
SHIP      & 371   & $59.84 \pm 11.13$ & 47.44  & SHIP      & 371   & $60.13 \pm 11.27$ & 42.05  \\
MESA      & 169   & $74.59 \pm 7.45$  & 46.15  & MESA      & 169   & $74.72 \pm 7.52$  & 46.15  \\
BLSA      & 107   & $68.47 \pm 9.96$  & 55.14  & BLSA      & 107   & $68.41 \pm 9.93$  & 55.14  \\
CARDIA    & 108   & $51.66 \pm 3.75$  & 56.48  & CARDIA    & 108   & $51.74 \pm 3.88$  & 56.48  \\
OASIS     & 57    & $70.41 \pm 8.73$  & 57.89  & OASIS     & 57    & $70.48 \pm 8.60$  & 57.89  \\
AIBL      & 55    & $72.60 \pm 5.99$  & 50.91  & AIBL      & 55    & $72.58 \pm 5.90$  & 50.91  \\
HANDLS    & 41    & $55.00 \pm 9.51$  & 53.66  & HANDLS    & 41    & $54.77 \pm 9.33$  & 51.22  \\
ADNI      & 28    & $73.87 \pm 5.70$  & 32.14  & ADNI      & 28    & $74.03 \pm 5.69$  & 32.14  \\
WHIMS     & 7     & $68.00 \pm 1.41$  & 100.00 & WHIMS     & 7     & $68.00 \pm 1.41$  & 100.00 \\
PreventAD & 7     & $67.26 \pm 5.85$  & 57.14  & PreventAD & 7     & $67.41 \pm 6.23$  & 57.14  \\
WRAP      & 2     & $70.00 \pm 0.40$  & 50.00  & WRAP      & 2     & $65.05 \pm 0.35$  & 100.00 \\
HCP       & 2     & $29.00 \pm 2.00$  & 0.00   & HCP       & 2     & $29.00 \pm 2.00$  & 0.00   \\
PENN      & 1     & $69.00 \pm 0.00$  & 100.00 & PENN      & 1     & $65.00 \pm 0.00$  & 100.00 \\
\midrule
\textbf{Total} & \textbf{3,006} & $\bm{65.49 \pm 9.04}$ & \textbf{40.52} &
\textbf{Total} & \textbf{3,006} & $\bm{65.55 \pm 9.06}$ & \textbf{39.85} \\
\bottomrule
\end{tabular}%
}
\end{table}
 
\begin{table}[H]
\centering
\small
\caption{Characteristics of the training and validation datasets for CU vs. Elevated BMI classification.}
\label{tab:dx_cn_bmi}
\resizebox{\textwidth}{!}{%
\begin{tabular}{l r c r @{\hspace{1.2cm}} l r c r}
\toprule
\multicolumn{4}{c}{\textit{CU}} &
\multicolumn{4}{c}{\textit{BMI}} \\
\cmidrule(lr){1-4}\cmidrule(lr){5-8}
\textbf{Study} & \textbf{n} & \textbf{Mean Age $\pm$ Std} & \textbf{(\%) F} &
\textbf{Study} & \textbf{n} & \textbf{Mean Age $\pm$ Std} & \textbf{(\%) F} \\
\midrule
UKBIOBANK & 1,702 & $62.17 \pm 7.17$  & 61.69  & UKBIOBANK & 1,702 & $62.16 \pm 7.18$  & 61.69  \\
CARDIA    & 132   & $50.22 \pm 3.94$  & 72.73  & CARDIA    & 132   & $50.16 \pm 3.86$  & 72.73  \\
HCP       & 122   & $37.25 \pm 13.59$ & 68.03  & HCP       & 122   & $37.27 \pm 13.59$ & 68.03  \\
SHIP      & 127   & $56.00 \pm 12.29$ & 63.78  & SHIP      & 127   & $56.00 \pm 12.29$ & 63.78  \\
MESA      & 106   & $70.70 \pm 5.76$  & 68.87  & MESA      & 106   & $70.61 \pm 5.94$  & 68.87  \\
ADNI      & 71    & $70.35 \pm 5.86$  & 70.42  & ADNI      & 71    & $70.38 \pm 5.89$  & 70.42  \\
OASIS     & 66    & $67.98 \pm 6.86$  & 69.70  & OASIS     & 66    & $67.86 \pm 6.99$  & 69.70  \\
BLSA      & 55    & $61.71 \pm 10.62$ & 67.27  & BLSA      & 55    & $61.89 \pm 10.68$ & 67.27  \\
HANDLS    & 47    & $52.03 \pm 8.62$  & 76.60  & HANDLS    & 47    & $51.92 \pm 9.01$  & 78.72  \\
AIBL      & 23    & $71.82 \pm 4.97$  & 69.57  & AIBL      & 23    & $71.87 \pm 4.97$  & 69.57  \\
PreventAD & 16    & $64.50 \pm 4.98$  & 75.00  & PreventAD & 16    & $64.46 \pm 5.04$  & 75.00  \\
BIOCARD   & 14    & $49.81 \pm 9.72$  & 71.43  & BIOCARD   & 13    & $50.81 \pm 9.40$  & 69.23  \\
WHIMS     & 8     & $68.00 \pm 2.87$  & 100.00 & WHIMS     & 8     & $68.00 \pm 2.87$  & 100.00 \\
WRAP      & 2     & $70.00 \pm 0.40$  & 50.00  & WRAP      & 2     & $70.90 \pm 0.40$  & 100.00 \\
PENN      & 1     & $70.20 \pm 0.00$  & 100.00 & PENN      & 1     & $70.91 \pm 0.00$  & 100.00 \\
\midrule
\textbf{Total} & \textbf{2,492} & $\bm{60.62 \pm 10.41}$ & \textbf{64.21} &
\textbf{Total} & \textbf{2,491} & $\bm{60.61 \pm 10.42}$ & \textbf{64.27} \\
\bottomrule
\end{tabular}%
}
\end{table}
 
\begin{table}[H]
\centering
\small
\caption{Characteristics of the training and validation datasets for smoker vs. non-smoker classification.}
\label{tab:dx_cn_smoking}
\resizebox{\textwidth}{!}{%
\begin{tabular}{l r c r @{\hspace{1.2cm}} l r c r}
\toprule
\multicolumn{4}{c}{\textit{Non-smoker}} &
\multicolumn{4}{c}{\textit{Smoker}} \\
\cmidrule(lr){1-4}\cmidrule(lr){5-8}
\textbf{Study} & \textbf{n} & \textbf{Mean Age $\pm$ Std} & \textbf{(\%) F} &
\textbf{Study} & \textbf{n} & \textbf{Mean Age $\pm$ Std} & \textbf{(\%) F} \\
\midrule
UKBIOBANK & 1,051 & $61.52 \pm 7.57$  & 45.58  & UKBIOBANK & 1,051 & $61.55 \pm 7.56$  & 45.58  \\
SHIP      & 680   & $52.71 \pm 13.72$ & 42.35  & SHIP      & 680   & $52.46 \pm 13.67$ & 61.62  \\
HCP       & 180   & $29.41 \pm 3.34$ & 43.89   & HCP       & 180   & $29.27 \pm 3.22$  & 43.89  \\
CARDIA    & 144   & $50.54 \pm 4.16$  & 48.61  & CARDIA    & 144   & $50.31 \pm 3.80$  & 48.61  \\
ADNI      & 95    & $74.97 \pm 4.69$  & 43.16  & ADNI      & 95    & $75.00 \pm 4.59$  & 43.16  \\
MESA      & 62    & $70.26 \pm 6.50$  & 53.23  & MESA      & 62    & $70.29 \pm 6.49$  & 53.23  \\
OASIS     & 44    & $65.75 \pm 8.83$  & 70.45  & OASIS     & 44    & $65.88 \pm 8.55$  & 70.45  \\
BLSA      & 30    & $62.90 \pm 12.08$ & 30.00  & BLSA      & 30    & $62.90 \pm 11.93$ & 30.00  \\
HANDLS    & 47    & $51.33 \pm 9.04$  & 55.32  & HANDLS    & 47    & $51.91 \pm 9.80$  & 48.94  \\
AIBL      & 22    & $70.21 \pm 7.55$  & 45.45  & AIBL      & 22    & $70.23 \pm 6.96$  & 45.45  \\
WHIMS     & 4     & $69.25 \pm 2.17$  & 100.00 & WHIMS     & 4     & $69.00 \pm 1.87$  & 100.00 \\
\midrule
\textbf{Total} & \textbf{2,359} & $\bm{56.62 \pm 13.64}$ & \textbf{45.36} &
\textbf{Total} & \textbf{2,359} & $\bm{56.55 \pm 13.67}$ & \textbf{50.78} \\
\bottomrule
\end{tabular}%
}
\end{table}
 
\begin{table}[H]
\centering
\small
\caption{Characteristics of the training and validation datasets for CU vs. Hypertension classification.}
\label{tab:dx_cn_hypertension}
\resizebox{\textwidth}{!}{%
\begin{tabular}{l r c r @{\hspace{1.2cm}} l r c r}
\toprule
\multicolumn{4}{c}{\textit{CU}} &
\multicolumn{4}{c}{\textit{Hypertension}} \\
\cmidrule(lr){1-4}\cmidrule(lr){5-8}
\textbf{Study} & \textbf{n} & \textbf{Mean Age $\pm$ Std} & \textbf{(\%) F} &
\textbf{Study} & \textbf{n} & \textbf{Mean Age $\pm$ Std} & \textbf{(\%) F} \\
\midrule
UKBIOBANK & 1,484 & $61.85 \pm 7.13$ & 58.22 & UKBIOBANK & 1,484 & $61.95 \pm 7.33$ & 64.35 \\
SHIP      & 82    & $51.96 \pm 11.53$ & 53.66 & SHIP      & 82    & $50.00 \pm 11.45$ & 71.95 \\
PreventAD & 9     & $66.87 \pm 6.47$  & 66.67 & PreventAD & 9     & $63.98 \pm 4.29$  & 100.00 \\
MESA      & 46    & $70.30 \pm 6.40$  & 47.83 & MESA      & 46    & $71.35 \pm 8.35$  & 52.17 \\
HANDLS    & 12    & $50.52 \pm 9.33$  & 50.00 & HANDLS    & 12    & $51.02 \pm 8.37$  & 83.33 \\
HCP       & 59    & $43.90 \pm 20.36$ & 35.59 & HCP       & 59    & $46.89 \pm 19.61$ & 71.19 \\
AIBL      & 16    & $74.28 \pm 5.67$  & 62.50 & AIBL      & 16    & $70.99 \pm 4.68$  & 50.00 \\
CARDIA    & 53    & $51.51 \pm 3.68$  & 49.06 & CARDIA    & 53    & $50.79 \pm 4.70$  & 49.06 \\
OASIS     & 38    & $69.58 \pm 7.05$  & 65.79 & OASIS     & 38    & $69.21 \pm 8.17$  & 60.53 \\
ADNI      & 35    & $70.66 \pm 7.13$  & 51.43 & ADNI      & 35    & $71.36 \pm 5.96$  & 42.86 \\
BLSA      & 47    & $71.64 \pm 8.81$  & 70.21 & BLSA      & 47    & $69.49 \pm 11.41$ & 42.55 \\
WHIMS     & 5     & $69.40 \pm 5.28$  & 100.00 & WHIMS    & 5     & $67.80 \pm 3.92$  & 100.00 \\
WRAP      & 1     & $70.50 \pm 0.00$  & 100.00 & WRAP     & 1     & $70.40 \pm 0.00$  & 100.00 \\
\midrule
\textbf{Total} & \textbf{1,887} & $\bm{61.42 \pm 9.56}$ & \textbf{57.29} &
\textbf{Total} & \textbf{1,887} & $\bm{61.42 \pm 9.63}$ & \textbf{63.43} \\
\bottomrule
\end{tabular}%
}
\end{table}

\subsection{Brain age gap differences across disease and risk-factor groups}

We evaluated whether the brain age gap (BAG, predicted minus chronological age) captures disease and risk-factor related effects. Using two brain age models, Brain2Vec-SVR and ViT-BA, we computed BAG in multiple disease-specific datasets and compared each disease or risk-factor group with its matched control group. A higher BAG in the affected groups would indicate that the models’ age predictions are sensitive to disease and risk-factor related changes in brain structure, beyond normal aging.

\begin{figure}[H]
    \begin{subfigure}[t]{0.48\linewidth}
      \includegraphics[width=\linewidth]{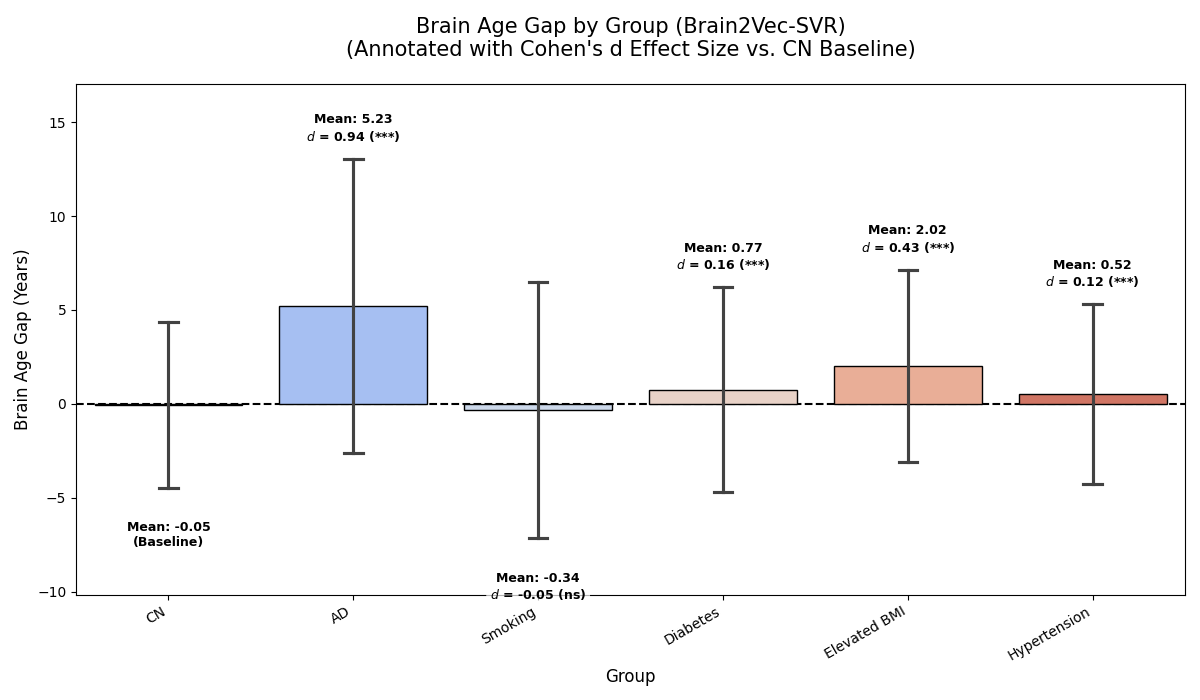}
      \caption{Brain2Vec-SVR models. For the inference, we ensemble predictions from all 5-fold SVR models.}
      \label{fig:brain2vec_svr_ba_gap}
    \end{subfigure}
    \hfill
    \begin{subfigure}[t]{0.48\linewidth}
      \includegraphics[width=\linewidth]{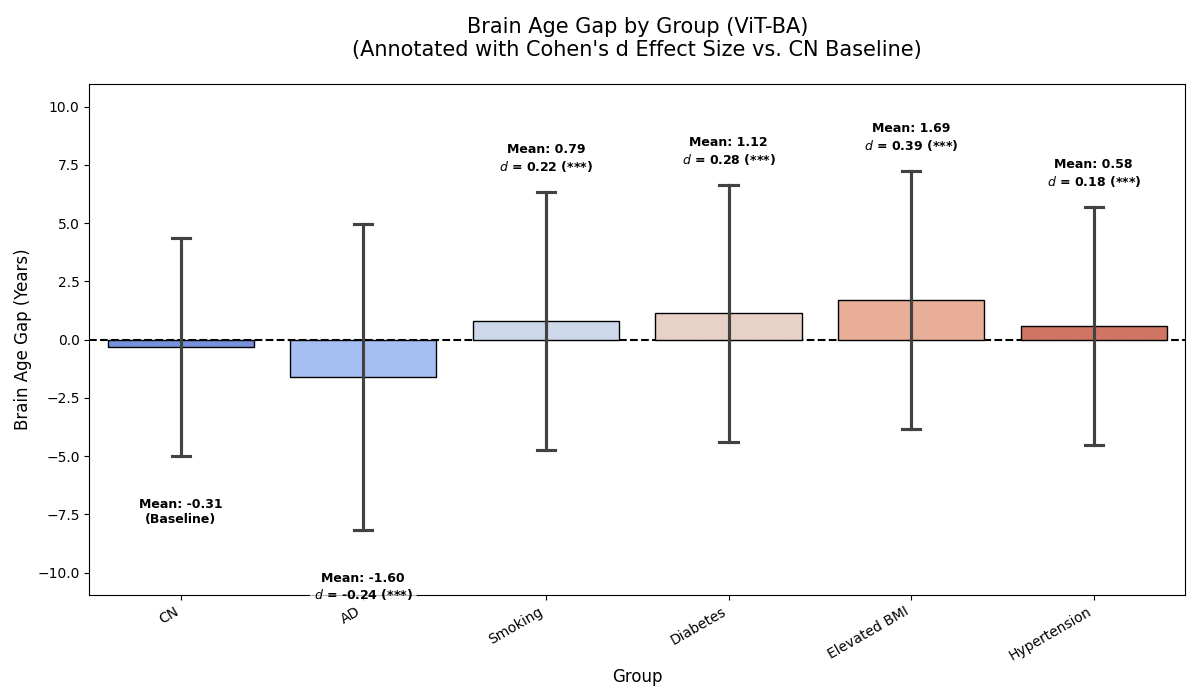}
      \caption{ViT-BA model.}
      \label{fig:vit_ba_gap}
    \end{subfigure}
    \caption{Brain age gap differences across disease and risk-factor groups.}
    \label{fig:ba_gap}
\end{figure}

\subsection{Brain2Vec classification performance across studies}
\begin{table}[H]
  \centering
  \caption{Disease classification performance by study using Brain2Vec embeddings as input features.}
  \label{tab:per_site_results}
  \resizebox{\linewidth}{!}{%
  \begin{tabular}{lcccccc}
    \toprule
    Study & n & Accuracy $\uparrow$ & AUC $\uparrow$ & F1 $\uparrow$ & Recall $\uparrow$ & Precision $\uparrow$ \\
    \midrule
    \multicolumn{7}{l}{\textit{CU vs AD}} \\
    \midrule
    ADNI & 837 & $0.85 \pm 0.01$ & $0.93 \pm 0.01$ & $0.85 \pm 0.01$ & $0.83 \pm 0.01$ & $0.86 \pm 0.01$ \\
    OASIS & 626 & $0.78 \pm 0.02$ & $0.86 \pm 0.02$ & $0.78 \pm 0.02$ & $0.78 \pm 0.02$ & $0.79 \pm 0.01$ \\
    AIBL & 184 & $0.86 \pm 0.00$ & $0.93 \pm 0.01$ & $0.85 \pm 0.01$ & $0.80 \pm 0.02$ & $0.90 \pm 0.01$ \\
    \midrule
    \multicolumn{7}{l}{\textit{CU vs Elevated BMI}} \\
    \midrule
    UKBIOBANK & 3404 & $0.84 \pm 0.01$ & $0.91 \pm 0.01$ & $0.84 \pm 0.01$ & $0.83 \pm 0.01$ & $0.84 \pm 0.00$ \\
    CARDIA & 264 & $0.72 \pm 0.01$ & $0.80 \pm 0.01$ & $0.71 \pm 0.01$ & $0.83 \pm 0.01$ & $0.84 \pm 0.00$ \\
    SHIP & 254 & $0.73 \pm 0.01$ & $0.81 \pm 0.01$ & $0.71 \pm 0.01$ & $0.68 \pm 0.02$ & $0.75 \pm 0.02$ \\
    \midrule
    \multicolumn{7}{l}{\textit{CU vs Diabetes}} \\
    \midrule
    UKBIOBANK & 4102 & $0.70 \pm 0.01$ & $0.76 \pm 0.01$ & $0.69 \pm 0.01$ & $0.67 \pm 0.01$ & $0.71 \pm 0.00$ \\
    SHIP & 742 & $0.65 \pm 0.01$ & $0.69 \pm 0.01$ & $0.61 \pm 0.01$ & $0.57 \pm 0.03$ & $0.67 \pm 0.02$ \\
    MESA & 338 & $0.60 \pm 0.01$ & $0.67 \pm 0.01$ & $0.67 \pm 0.01$ & $0.81 \pm 0.05$ & $0.57 \pm 0.01$ \\
    \midrule
    \multicolumn{7}{l}{\textit{CU vs Smoking}} \\
    \midrule
    UKBIOBANK & 2102 & $0.69 \pm 0.01$ & $0.76 \pm 0.01$ & $0.71 \pm 0.01$ & $0.73 \pm 0.02$ & $0.68 \pm 0.01$ \\
    SHIP & 1360 & $0.65 \pm 0.01$ & $0.70 \pm 0.01$ & $0.66 \pm 0.01$ & $0.68 \pm 0.02$ & $0.64 \pm 0.01$ \\
    HCP & 360 & $0.61 \pm 0.02$ & $0.67 \pm 0.01$ & $0.61 \pm 0.04$ & $0.61 \pm 0.08$ & $0.61 \pm 0.02$ \\
    \midrule
    \multicolumn{7}{l}{\textit{CU vs Hypertension}} \\
    \midrule
    UKBIOBANK & 17797 & $0.63 \pm 0.00$ & $0.67 \pm 0.00$ & $0.65 \pm 0.01$ & $0.69 \pm 0.02$ & $0.61 \pm 0.01$ \\
    SHIP & 978 & $0.58 \pm 0.02$ & $0.62 \pm 0.01$ & $0.60 \pm 0.04$ & $0.63 \pm 0.10$ & $0.58 \pm 0.02$ \\
    HCP & 708 & $0.53 \pm 0.01$ & $0.55 \pm 0.01$ & $0.58 \pm 0.07$ & $0.66 \pm 0.16$ & $0.53 \pm 0.01$ \\
    \bottomrule
  \end{tabular}
  }
\end{table}

\subsection{Study-related variation in Brain2Vec embeddings for different disease and risk-factor groups}

To assess whether Brain2Vec embeddings encode acquisition site in addition to anatomy, we projected the embeddings of each disease and risk-factor group onto their first two principal components and colored each scan by site.

\begin{figure}[H]
    \begin{subfigure}[t]{0.32\linewidth}
      \includegraphics[width=\linewidth]{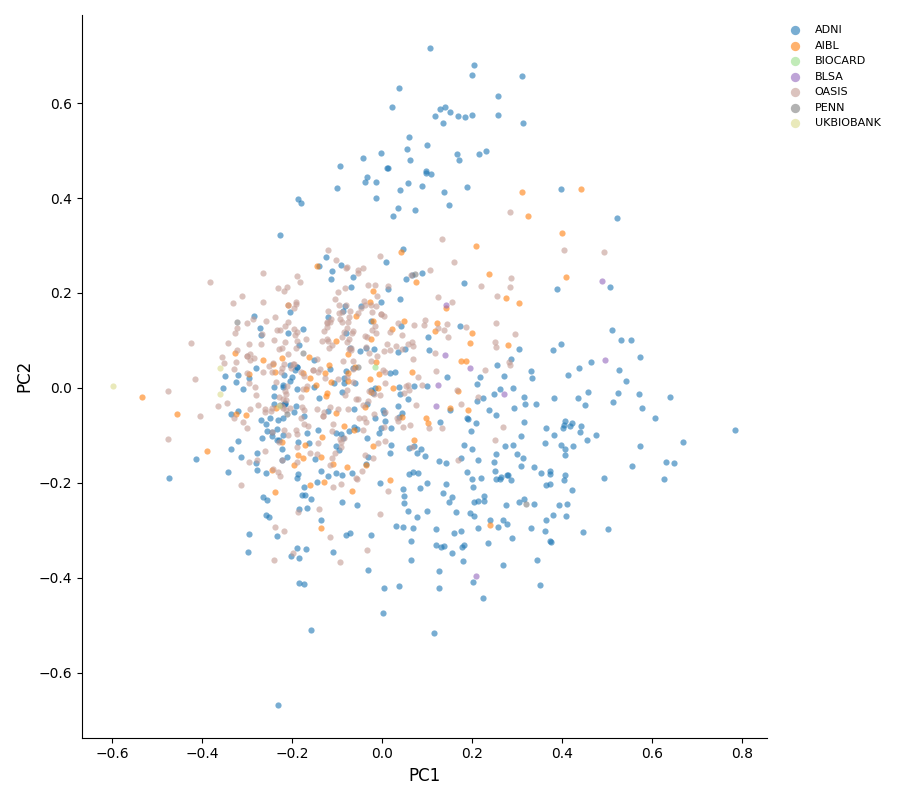}
      \caption{PCA on AD.}
      \label{fig:pca_ad}
    \end{subfigure}
    \hfill
    \begin{subfigure}[t]{0.32\linewidth}
      \includegraphics[width=\linewidth]{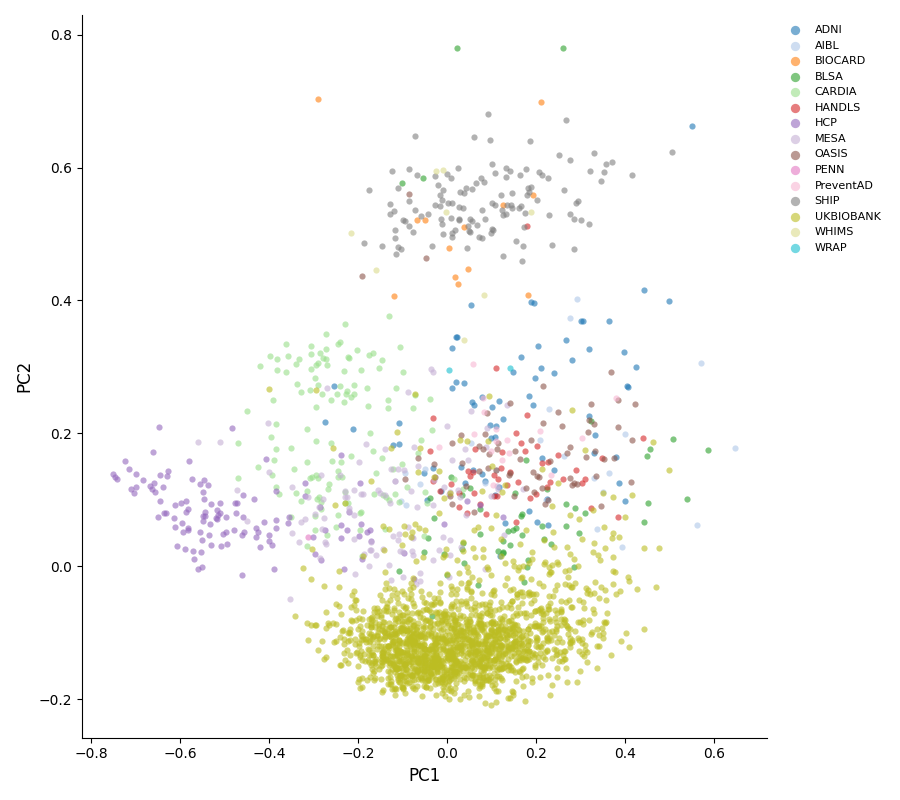}
      \caption{PCA on Elevated BMI.}
      \label{fig:pca_bmi}
    \end{subfigure}
    \hfill
    \begin{subfigure}[t]{0.32\linewidth}
      \includegraphics[width=\linewidth]{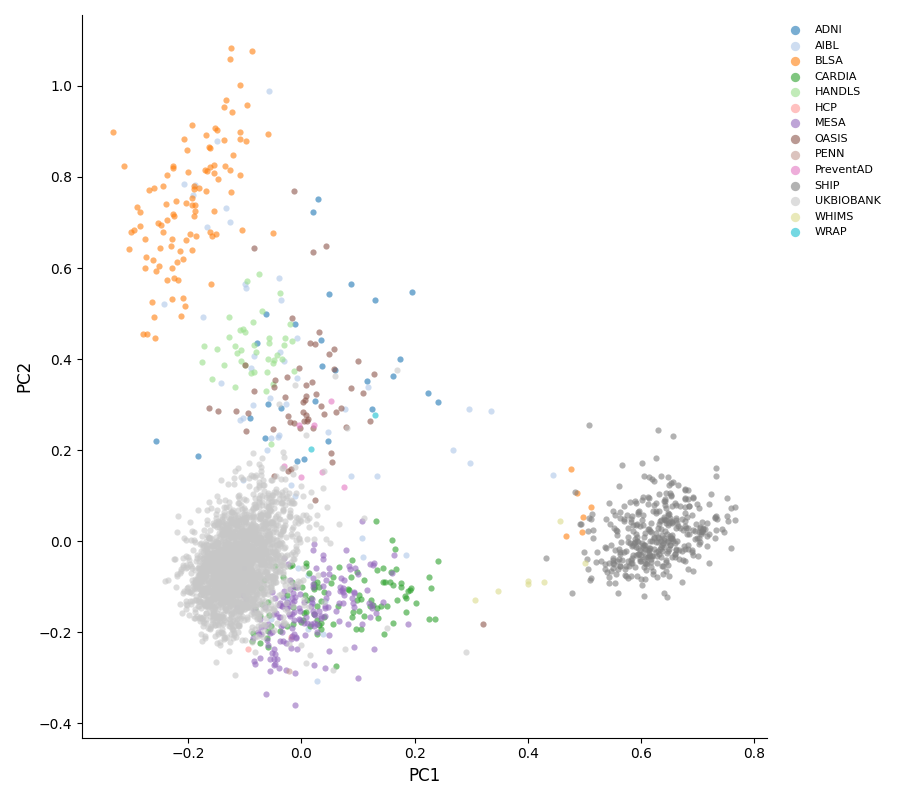}
      \caption{PCA on Diabetes.}
      \label{fig:pca_diabetes}
    \end{subfigure}

    \vspace{0.5em}
    \hspace*{0.17\linewidth}
    \begin{subfigure}[t]{0.32\linewidth}
      \includegraphics[width=\linewidth]{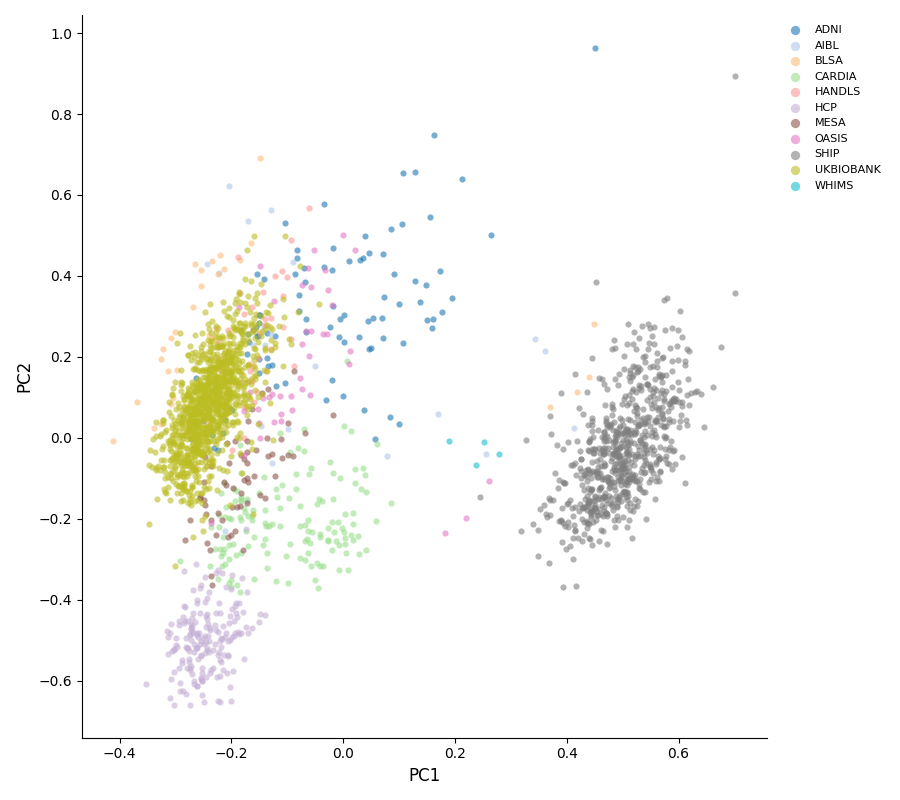}
      \caption{PCA on Smoking.}
      \label{fig:pca_smoking}
    \end{subfigure}
    \hfill
    \begin{subfigure}[t]{0.32\linewidth}
      \includegraphics[width=\linewidth]{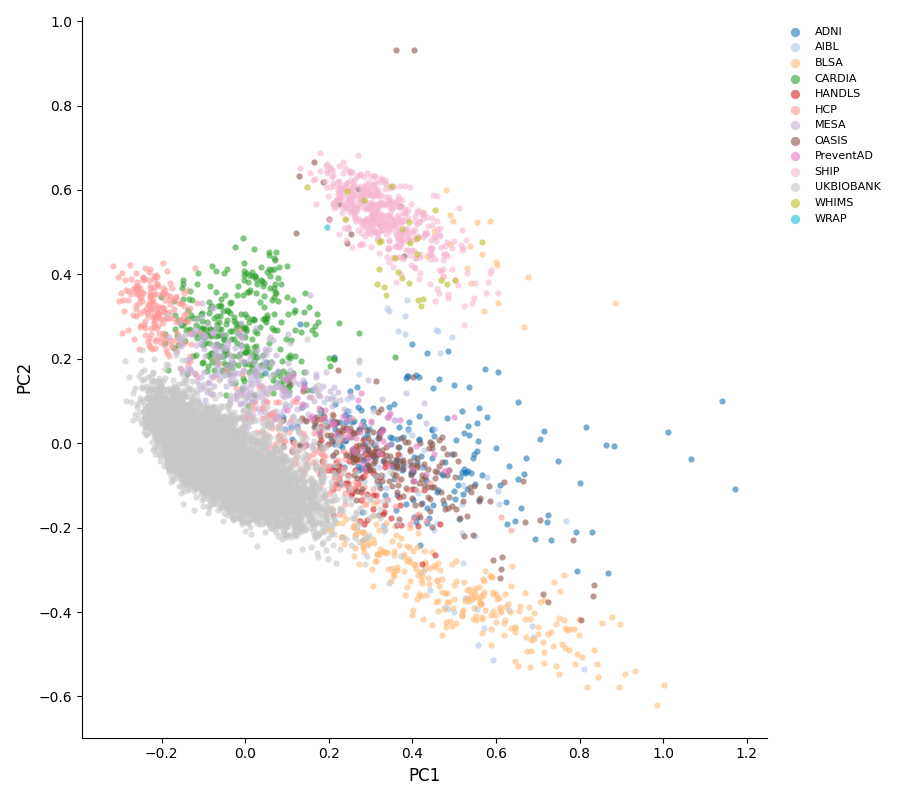}
      \caption{PCA on Hypertension.}
      \label{fig:pca_hypertension}
    \end{subfigure}
    \hspace*{0.17\linewidth}

    \caption{Principal component analysis of Brain2Vec embeddings for samples in different disease and risk-factor groups, colored by acquisition site.}
    \label{fig:pca_batch_effects}
\end{figure}

\subsection{Selection of Brain2Vec features for latent manipulation}

To construct the subset of manipulated features $S$ used in Section~\ref{sec:brain2vec_embedding_analysis}, we identified Brain2Vec latent dimensions that are strongly associated with specific DLMUSE ROIs and have a consistent, monotonic effect on those regions under manipulation. For each Brain2Vec dimension, we computed (i) the mean difference between CU and AD subjects and (ii) its Pearson correlation with the best-matching DLMUSE ROI. Based on this ranking, we selected the 22 dimensions with the largest and most consistent effects (supplementary Table~\ref{tab:brain2vec_important_features}).

We restricted this latent manipulation analysis to the CU vs.\ AD task because it shows the largest anatomical differences between groups among the classification tasks we studied. We repeated the same selection procedure for other tasks in our cohort, including elevated BMI, smoking, diabetes, and hypertension, but did not observe features with comparably strong effects. This suggests that the latent directions identified for CU vs.\ AD reflect a stronger, more anatomically coherent signal than those for tasks with subtler or more heterogeneous structural correlates.

\begin{table}[!tbp]
  \centering
  \caption{Brain2Vec dimensions selected for latent manipulation and their best-matching DLMUSE ROIs}
  \label{tab:brain2vec_important_features}
  \resizebox{\linewidth}{!}{%
  \begin{tabular}{lcccc}
    \toprule
    Brain2Vec dim. & Mean difference & Cohen's $d$ & DLMUSE ROI & Pearson $r$ \\
    \midrule
    497 & $2.90\times10^{-5}$  & $0.8365$  & 509 (Ventricles) & $0.6624$  \\
    110 & $-2.87\times10^{-2}$ & $-0.7359$ & 509 (Ventricles) & $-0.5890$ \\
    327 & $2.74\times10^{-2}$  & $0.7299$  & 509 (Ventricles) & $0.4885$  \\
    175 & $-1.63\times10^{-2}$ & $-0.7027$ & 509 (Ventricles) & $-0.5720$ \\
    693 & $-9.09\times10^{-3}$ & $-0.6911$ & 509 (Ventricles) & $-0.5170$ \\
    691 & $-1.95\times10^{-2}$ & $-0.6884$ & 509 (Ventricles) & $-0.4771$ \\
    451 & $2.81\times10^{-2}$  & $0.6811$  & 4 (3rd Ventricle) & $0.6381$  \\
    230 & $1.40\times10^{-2}$  & $0.6756$  & 4 (3rd Ventricle) & $0.4295$  \\
    583 & $2.07\times10^{-2}$  & $0.6637$  & 509 (Ventricles) & $0.6794$  \\
    249 & $-2.28\times10^{-2}$ & $-0.6602$ & 4 (3rd Ventricle) & $-0.5147$ \\
    521 & $-3.69\times10^{-4}$ & $-0.6540$ & 4 (3rd Ventricle) & $-0.4167$ \\
    427 & $1.52\times10^{-2}$  & $0.6466$  & 4 (3rd Ventricle) & $0.4494$  \\
    188 & $1.87\times10^{-2}$  & $0.6416$  & 4 (3rd Ventricle) & $0.5080$  \\
    426 & $1.90\times10^{-2}$  & $0.6399$  & 509 (Ventricles) & $0.4882$  \\
    120 & $-3.3\times10^{-5}$ & $-0.6355$ & 4 (3rd Ventricle) & $-0.5458$ \\
    431 & $5.29\times10^{-3}$  & $0.6327$  & 4 (3rd Ventricle) & $0.5534$  \\
    193 & $5.21\times10^{-2}$  & $0.6245$  & 509 (Ventricles) & $0.5414$  \\
    357 & $1.71\times10^{-2}$  & $0.6171$  & 509 (Ventricles) & $0.4215$  \\
    206 & $1.30\times10^{-5}$  & $0.6017$  & 509 (Ventricles) & $0.6608$  \\
    183 & $5.34\times10^{-3}$  & $0.5969$  & 4 (3rd Ventricle) & $0.4961$  \\
    201 & $-3.00\times10^{-6}$ & $-0.5930$ & 4 (3rd Ventricle) & $-0.5834$ \\
    491 & $-6.28\times10^{-4}$ & $-0.5909$ & 509 (Ventricles) & $-0.5279$ \\
    \bottomrule
  \end{tabular}
  }
\end{table}

\subsection{Example self-attention maps for CU, MCI, and AD participants}
\label{sec:self_attention}

\begin{figure}[H]
\centering
\includegraphics[width=\textwidth]{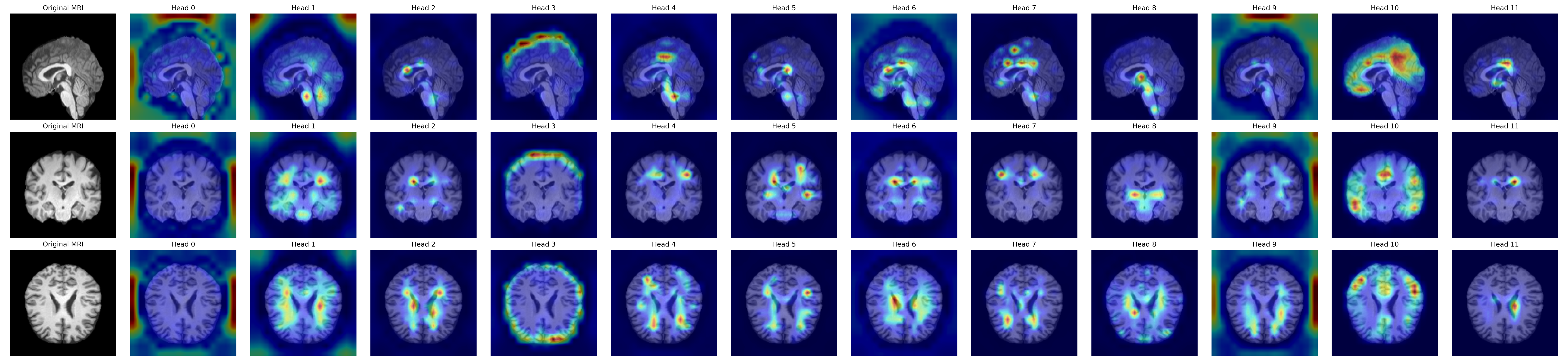}
\caption{Self-attention map of a cognitively unimpaired patient}
\label{fig:cn_self_attention}
\end{figure}

\begin{figure}[H]
\centering
\includegraphics[width=\textwidth]{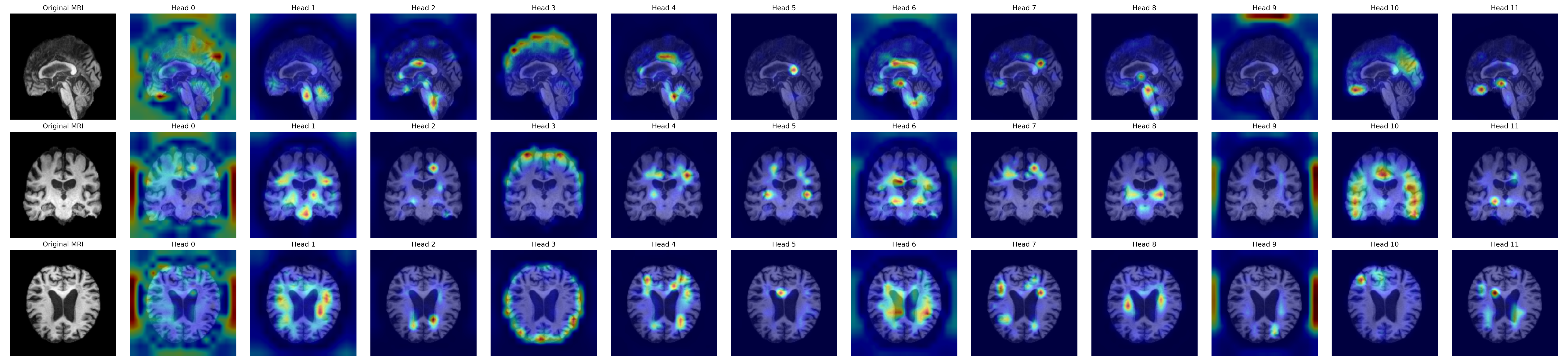}
\caption{Self attention map of a participant with mild cognitive impairment}
\label{fig:mci_self_attention}
\end{figure}

\begin{figure}[H]
\centering
\includegraphics[width=\textwidth]{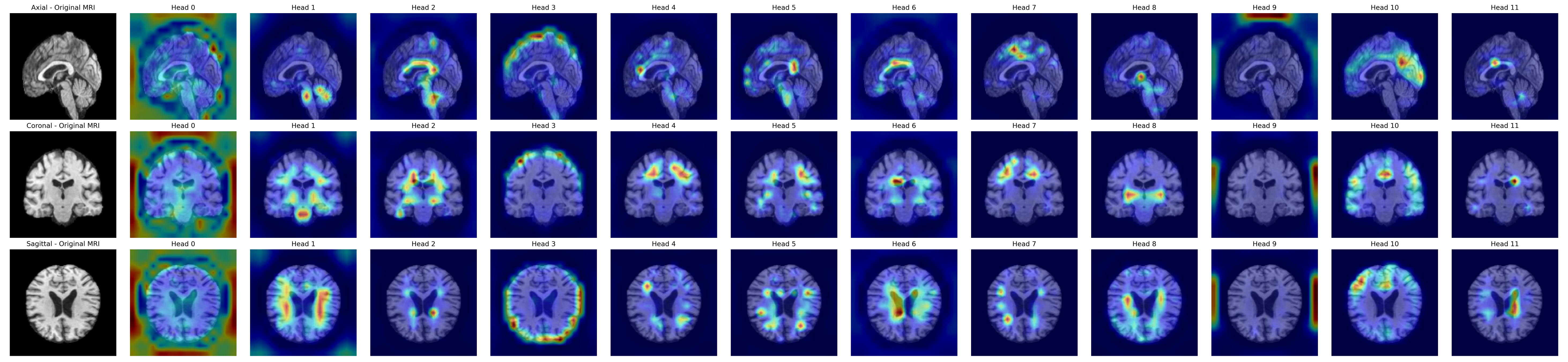}
\caption{Self-attention map of a participant with Alzheimer's disease}
\label{fig:ad_self_attention}
\end{figure}

\subsection{Manipulation of latent Brain2Vec embeddings}
\label{sec:embedding_manipulation}

\begin{figure}[H]
    \includegraphics[width=\linewidth]{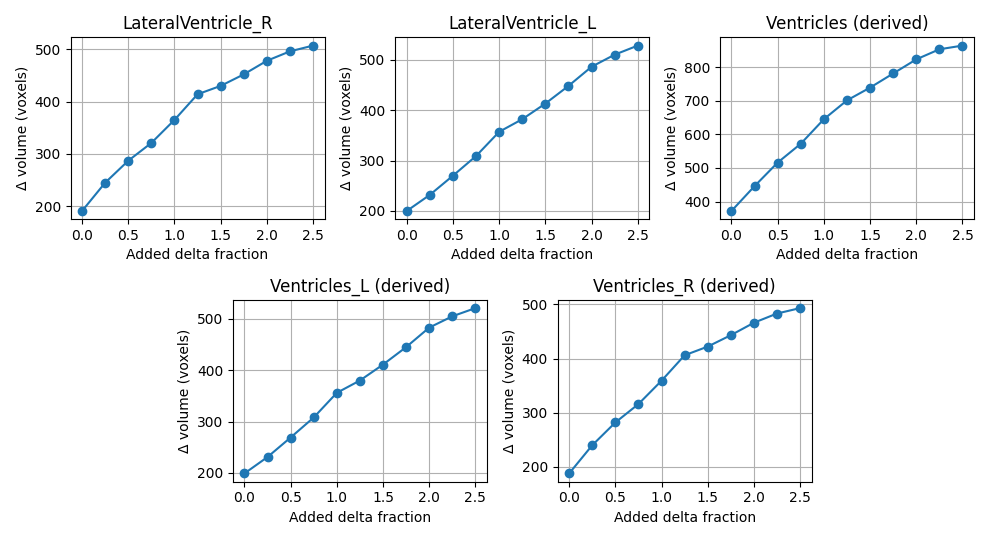}
    \caption{Effect on ventricles: we observe a monotonic increase of ventricle regions which is clinically expected, as ventricular enlargement is a well-established radiological marker of neurodegeneration and disease progression in Alzheimer's disease \citep{Nestor2008VentricularEA}.}
    \label{fig:ventricles_brain2vec_effect}
\end{figure}

\begin{figure}[H]
    \includegraphics[width=\linewidth]{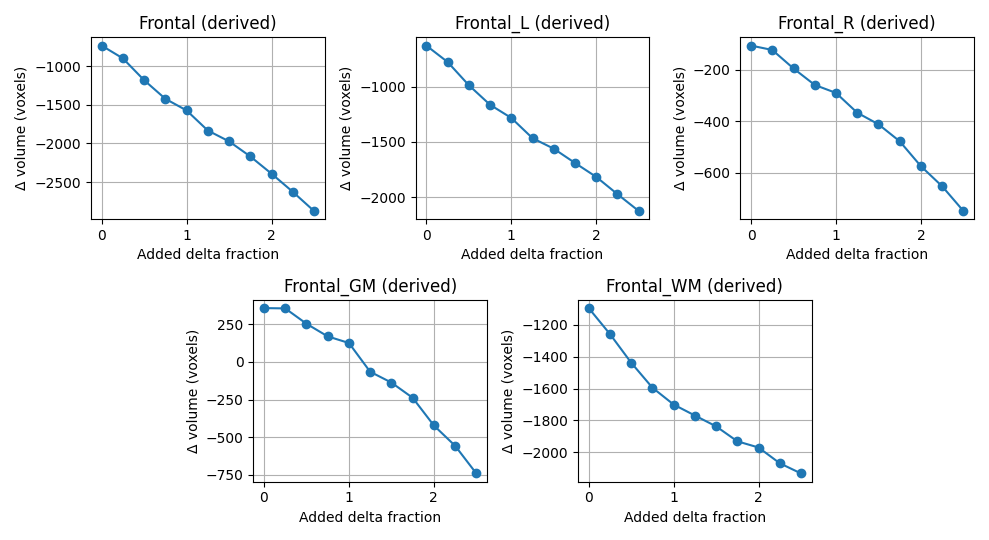}
    \caption{Effect on frontal regions: we observe a monotonic decrease of frontal regions which is consistent with well-documented patterns of medial and lateral frontal lobe atrophy in AD patients ~\citep{WHITWELL2012868, 10.1093/braincomms/fcac109}.}
    \label{fig:frontal_brain2vec_effect}
\end{figure}

\begin{figure}[H]
    \includegraphics[width=0.9\linewidth]{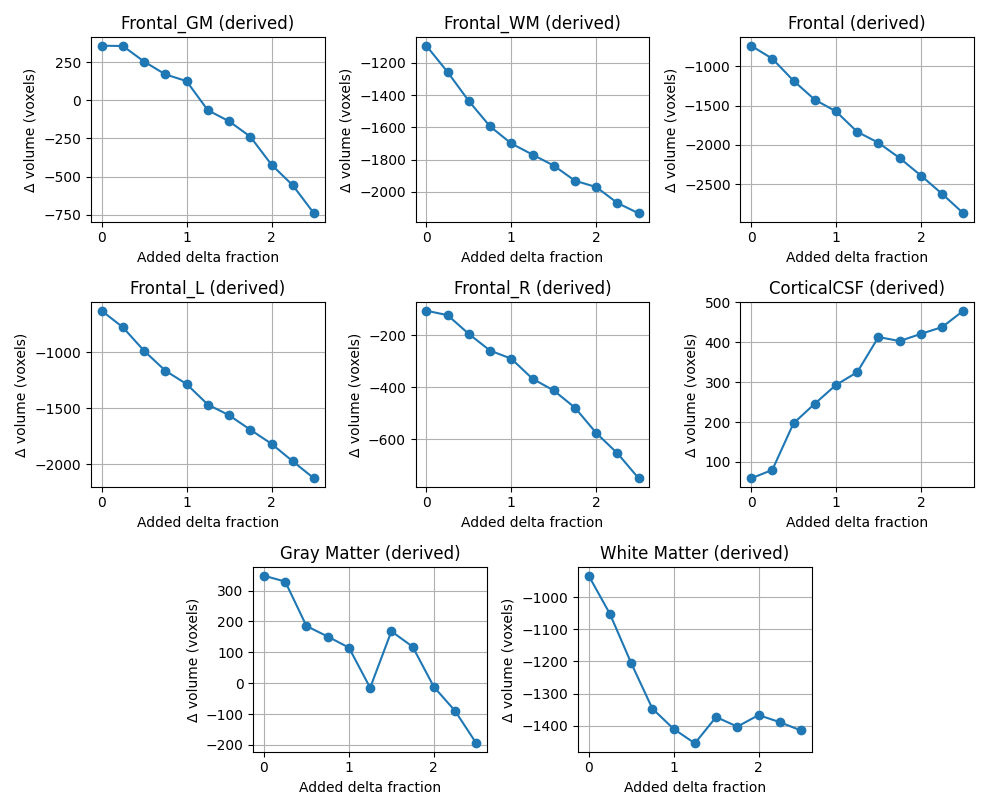}
    \caption{Effects on GM: we observe a similar trend on frontal regions as well as decreased GM and WM volumes, accompanied by an increase in cortical CSF. This pattern is consistent with enlargement of CSF spaces as brain tissue is lost, and is validated by prior volumetric findings in Alzheimer's disease~\citep{GUO2010146}.}
    \label{fig:gm_wm_brain2vec_effect}
\end{figure}

\subsection{Brain2Vec masking and reconstruction examples}

\begin{figure}[H]
    \centering
    \setlength{\subfigheight}{11.0cm}
    \begin{subfigure}[t]{0.23\linewidth}
    \centering
    \makebox[\linewidth]{\includegraphics[height=\subfigheight]{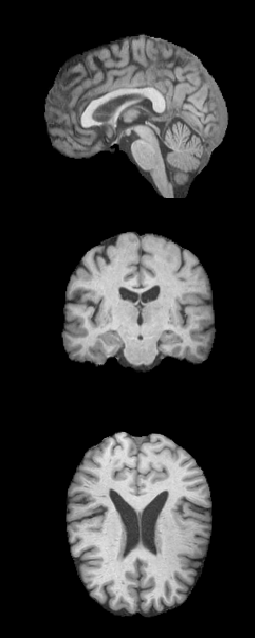}}
    \caption{Original input}
  \end{subfigure}%
  \begin{subfigure}[t]{0.23\linewidth}
    \centering
    \makebox[\linewidth]{\includegraphics[height=\subfigheight]{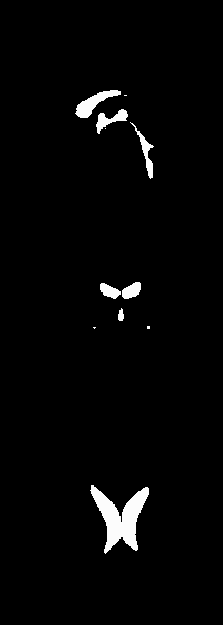}}
    \caption{Ventricles region mask}
  \end{subfigure}%
  \begin{subfigure}[t]{0.23\linewidth}
    \centering
    \makebox[\linewidth]{\includegraphics[height=\subfigheight]{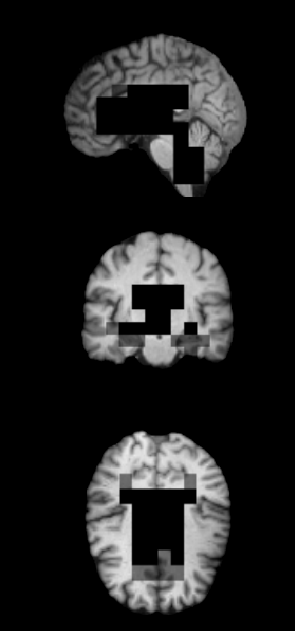}}
    \caption{Ventricles masking for Brain2Vec}
  \end{subfigure}%
  \begin{subfigure}[t]{0.23\linewidth}
    \centering
    \makebox[\linewidth]{\includegraphics[height=\subfigheight]{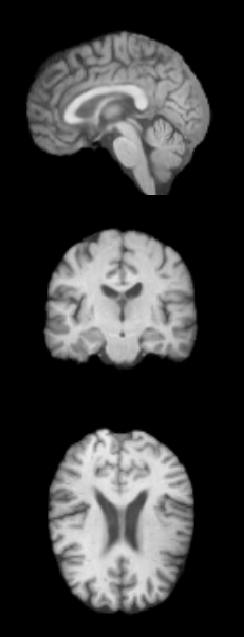}}
    \caption{Ventricles reconstruction with Brain2Vec}
  \end{subfigure}
  \caption{Example of ventricles region masking and reconstruction with Brain2Vec.}
  \label{fig:masking_ventricles}
\end{figure}

\begin{figure}[H]
    \centering
    \setlength{\subfigheight}{11.0cm}
    \begin{subfigure}[t]{0.23\linewidth}
    \centering
    \makebox[\linewidth]{\includegraphics[height=\subfigheight]{images/brain2vec_examples/brain2vec_latent_embeddings_analysis_scan_used.png}}
    \caption{Original input}
  \end{subfigure}%
  \begin{subfigure}[t]{0.23\linewidth}
    \centering
    \makebox[\linewidth]{\includegraphics[height=\subfigheight]{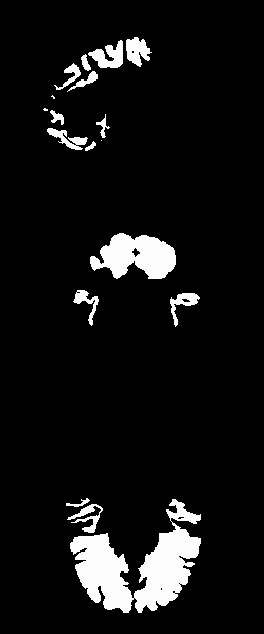}}
    \caption{Frontal region mask}
  \end{subfigure}%
  \begin{subfigure}[t]{0.23\linewidth}
    \centering
    \makebox[\linewidth]{\includegraphics[height=\subfigheight]{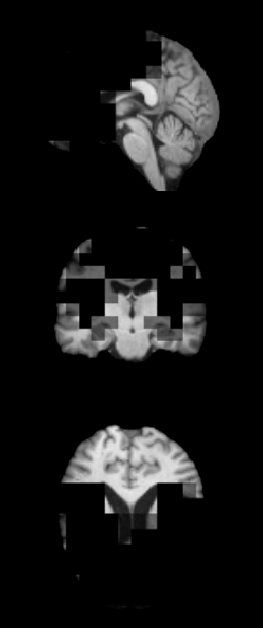}}
    \caption{Frontal masking for Brain2Vec}
  \end{subfigure}%
  \begin{subfigure}[t]{0.23\linewidth}
    \centering
    \makebox[\linewidth]{\includegraphics[height=\subfigheight]{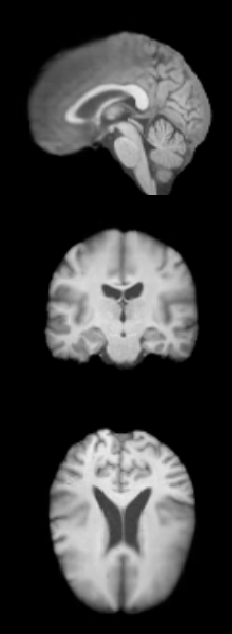}}
    \caption{Frontal reconstruction with Brain2Vec}
  \end{subfigure}
  \caption{Example of frontal region masking and reconstruction with Brain2Vec.}
  \label{fig:masking_frontal}
\end{figure}

\begin{figure}[H]
    \centering
    \setlength{\subfigheight}{11.0cm}
    \begin{subfigure}[t]{0.23\linewidth}
    \centering
    \makebox[\linewidth]{\includegraphics[height=\subfigheight]{images/brain2vec_examples/brain2vec_latent_embeddings_analysis_scan_used.png}}
    \caption{Original input}
  \end{subfigure}%
  \begin{subfigure}[t]{0.23\linewidth}
    \centering
    \makebox[\linewidth]{\includegraphics[height=\subfigheight]{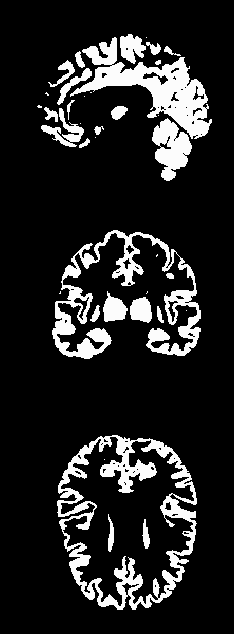}}
    \caption{GM region mask}
  \end{subfigure}%
  \begin{subfigure}[t]{0.23\linewidth}
    \centering
    \makebox[\linewidth]{\includegraphics[height=\subfigheight]{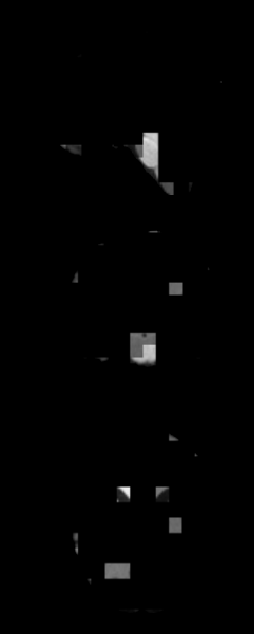}}
    \caption{GM masking for Brain2Vec}
  \end{subfigure}%
  \begin{subfigure}[t]{0.23\linewidth}
    \centering
    \makebox[\linewidth]{\includegraphics[height=\subfigheight]{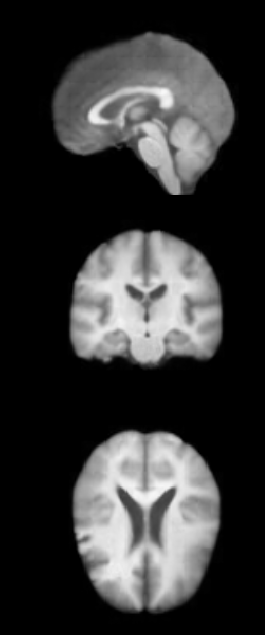}}
    \caption{GM reconstruction with Brain2Vec}
  \end{subfigure}
  \caption{Example of gray matter region masking and reconstruction with Brain2Vec.}
  \label{fig:masking_gm}
\end{figure}

\end{document}